\documentclass[essd, preprint]{copernicus}

\usepackage{tabularx}
\usepackage{cancel}
\usepackage{multirow}
\usepackage{supertabular}
\usepackage{algorithmic}
\usepackage{algorithm}
\usepackage{amsthm}
\usepackage{float}
\usepackage{subfig}
\usepackage{rotating}
\usepackage{lineno}
\nolinenumbers
\usepackage[table]{xcolor}

\begin{document}

\title{\textsc{Bright}: A globally distributed multimodal building damage assessment dataset with very-high-resolution for all-weather disaster response}


\Author[1,2]{Hongruixuan}{Chen} 
\Author[1,2]{Jian}{Song}
\Author[3]{Olivier}{Dietrich}
\Author[2]{Clifford}{Broni-Bediako}
\Author[1,2]{Weihao}{Xuan}
\Author[1]{Junjue}{Wang}
\Author[1]{Xinlei}{Shao}
\Author[1]{Yimin}{Wei}
\Author[2]{Junshi}{Xia}
\Author[4]{Cuiling}{Lan}
\Author[3]{Konrad}{Schindler}
\Author[1,2][yokoya@k.u-tokyo.ac.jp]{Naoto}{Yokoya} 

\affil[1]{Graduate School of Frontier Sciences, The University of Tokyo, Chiba, Japan}
\affil[2]{RIKEN Center for Advanced Intelligence Project (AIP), RIKEN, Tokyo, Japan}
\affil[3]{Department of Photogrammetry and Remote Sensing, ETH Zürich, Zürich, Switzerland}
\affil[4]{Microsoft Research Asia, Beijing, China}




\runningtitle{\textsc{Bright}: A globally distributed MM BDA dataset with VHR for all-weather disaster response}

\runningauthor{H. Chen et al.}

\received{}
\pubdiscuss{} 
\revised{}
\accepted{}
\published{}


\firstpage{1}

\maketitle

\begin{abstract}
Disaster events occur around the world and cause significant damage to human life and property. Earth observation (EO) data enables rapid and comprehensive building damage assessment, an essential capability crucial in the aftermath of a disaster to reduce human casualties and inform disaster relief efforts. Recent research focuses on developing artificial intelligence (AI) models to accurately map unseen disaster events, mostly using optical EO data. These solutions based on optical data are limited to clear skies and daylight hours, preventing a prompt response to disasters. Integrating multimodal EO data, particularly combining optical and synthetic aperture radar (SAR) imagery, makes it possible to provide all-weather, day-and-night disaster responses. Despite this potential, the lack of suitable benchmark datasets has constrained the development of robust multimodal AI models. In this paper, we present a \underline{B}uilding damage assessment dataset using ve\underline{R}y-h\underline{IGH}-resolu\underline{T}ion optical and SAR imagery (\textsc{Bright}) to support AI-based all-weather disaster response. To the best of our knowledge, \textsc{Bright} is the first open-access, globally distributed, event-diverse multimodal dataset specifically curated to support AI-based disaster response. It covers five types of natural disasters and two types of human-made disasters across 14 regions worldwide, focusing on developing countries where external assistance is most needed. The dataset's optical and SAR images with spatial resolutions between 0.3 and 1 meters provide detailed representations of individual buildings, making it ideal for precise damage assessment. We train seven advanced AI models on \textsc{Bright} to validate transferability and robustness. Beyond that, it also serves as a challenging benchmark for a variety of tasks in real-world disaster scenarios, including unsupervised domain adaptation, semi-supervised learning, unsupervised multimodal change detection, and unsupervised multimodal image matching. The experimental results serve as baselines to inspire future research and model development. The dataset \citep{Chen2025BRIGHT}, along with the code and pretrained models, is available at \textcolor{magenta}{\url{https://github.com/ChenHongruixuan/BRIGHT}} and will be updated as and when a new disaster data is available. \textsc{Bright} also serves as the official dataset for the 2025 IEEE GRSS Data Fusion Contest Track II. We hope that this effort will promote the development of AI-driven methods in support of people in disaster-affected areas. 

\end{abstract}


\introduction  
A disaster is defined as a severe disruption in the functioning of a community or society due to the interaction between a hazard event and the conditions of exposure, vulnerability and capacity resulting in human, material, economic or environmental losses and impacts \citep{Ge2020review}. According to the United Nations Office for Disaster Risk Reduction (UNDRR), between 1998 and 2017, natural disasters such as earthquakes, storms, and floods affected approximately 4.4 billion people and caused 1.3 million deaths. These disasters have also resulted in economic losses of 2,647 billion United States dollars (USD) in disaster-affected countries \citep{UNDRR2018EconomicLosses}. The threat of disasters is likely to increase due to global urbanization \citep{Kreibich2022, Moroz2024Urban}. Rapid and comprehensive damage assessment is crucial in the aftermath of a disaster to make informed and effective rescue decisions that minimize losses and impacts. Building damage assessment aims to provide information, including the area and amount of damage, the rate of collapsed buildings, and the type of damage each building has incurred. This information is critical in the early stages of a disaster, as the distribution of damaged buildings is closely related to life-saving efforts in an emergency response \citep{Xie2016Crowdsourcing, ADRIANO2021132}. Conducting field surveys after a disaster can be difficult and dangerous, especially when transportation and communication systems are disrupted, making efficient on-site assessments challenging. Earth observation (EO) provides a safe and efficient way to obtain information on building damage in disaster areas due to its wide field of view and contactless operation.

\par The EO technologies commonly used for assessing building damage after disasters are optical and synthetic aperture radar (SAR). Optical imagery is a primary source for building damage assessment because of its intuitive and easy-to-interpret nature. For example, moderate-resolution optical data from the Landsat series and Sentinel-2 have been used to assess building damage \citep{Yusuf2001Damage, Fan2019Estimating, Febri2022Sentinel}. Landsat and Sentinel-2 data are limited in spatial resolution and only provide broad approximations of affected areas, which lack precision for specific buildings, crucial for timely rescue. The new generation of very high-resolution (VHR) optical sensors, such as IKONOS and WorldView, provides EO data with spatial resolutions of a meter or less, enabling finer assessments at the level of individual buildings \citep{Freire2014Introducing}. These data have been used successfully in building damage assessment \citep{Yamazaki2007Remote, Tong2012Building, Freire2014Introducing}. 

\begin{figure*}[!t]
    \centering
    \includegraphics[width=6.95in]{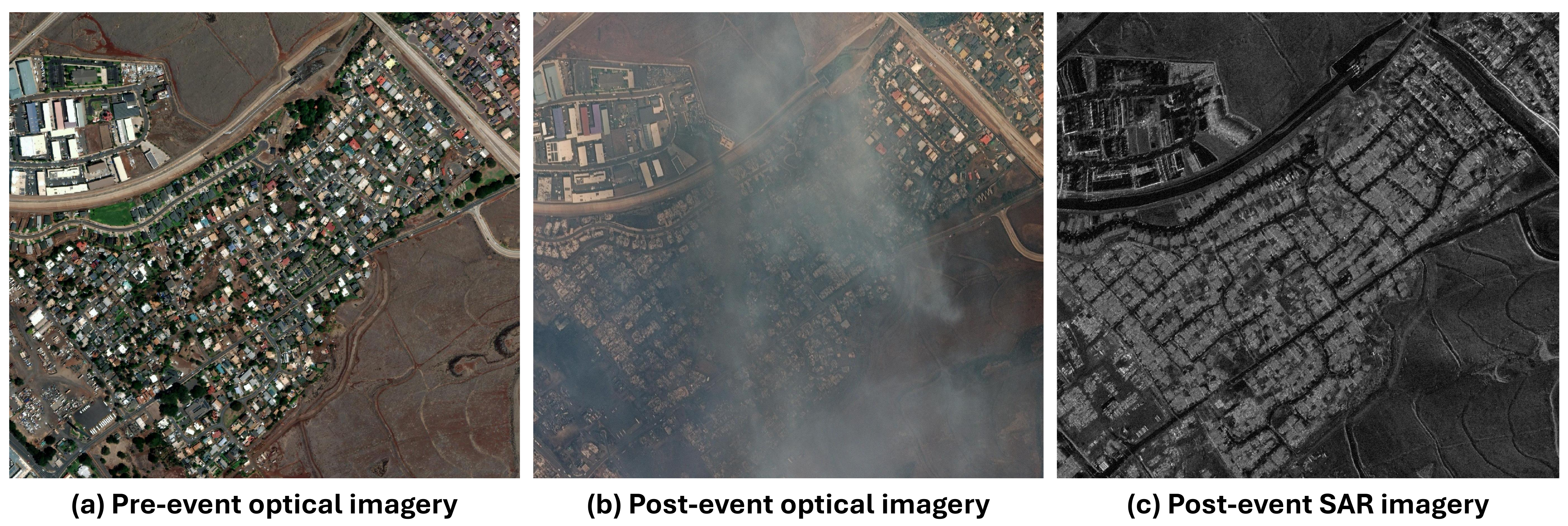}
   \caption{An example of the wildfire occurring in Maui, Hawaii, USA, August 2023. (a) Pre-event optical imagery (© Maxar). (b) Post-event optical image (© Maxar) with land-cover features obscured by wildfire smoke. (c) Post-event SAR imagery (© Capella Space) unaffected by smoke, showing the disaster area. }

  \label{fig:multimodal_image_case}
\end{figure*}

\par While accurate building damage maps can be obtained by visual interpretation of optical images by human experts, this process is time-consuming and labor-intensive for large-scale rapid assessments. In addition, it requires trained professionals. Therefore, recent studies have focused on developing automated methods for rapid building damage mapping \citep{Tong2012Building, Xie2016Crowdsourcing, Gupta_2019_CVPR_Workshops, ZHENG2021Building}. Among these, machine learning (ML) and deep learning (DL) techniques have significantly improved efficiency and accuracy in building damage assessment. Earlier work focused on a single disaster event with labels annotated for a specific disaster area to train a model. This model is then used to generate building damage maps for the same event \citep{Xie2016Crowdsourcing, Xia2023Deep}. However, since training data were limited to a few building types, damage patterns, and background land cover distributions, the resulting models mostly lack generalizability and struggle to produce accurate building damage maps for new disaster events, which limits their practical use. Recent large-scale benchmark datasets, for example, the xBD dataset \citep{Gupta_2019_CVPR_Workshops} containing different types of disaster scenarios and damages, have made it possible to adopt DL models to quickly and accurately map building damages after a newly occurred, previously unseen disaster \citep{ZHENG2021Building, Chen2022Dual, Shen2022BDANet, Kaur2023Large, Guo2024SAAN, Wang2024PCDASNet, Chen2024ChangeMamba}. For example, \cite{ZHENG2021Building} trained DL models on the xBD dataset and applied them to map the damage to buildings in two unseen human-made disaster events. These studies have demonstrated the effectiveness of DL models for building damage mapping. 

\par The optical EO technology uses a passive sensing technique, which requires solar illumination and cloud-free weather conditions. This severely limits the application of optical images in an emergency tool for all-weather disaster response \citep{ADRIANO2021132}. In contrast, SAR sensors employ active illumination with longer microwaves and can acquire images in adverse weather conditions, offering great potential for all-weather disaster response. Most disaster events, especially wildfires, floods, and storms, are often accompanied by less-than-ideal imaging conditions. For example, Figure \ref{fig:multimodal_image_case} shows EO imagery captured for a wildfire event that occurred in August 2023 in Hawaii, USA. The post-event optical image shown in Figure \ref{fig:multimodal_image_case}-(b) does not provide clear surface information due to the effects of the wildfire smoke. However, the SAR image illustrated in Figure \ref{fig:multimodal_image_case}-(c) is not affected by smoke and clearly shows the buildings damaged by the wildfire. 

\par Due to the advantages of SAR imagery, various SAR-based methods have been proposed for building damage assessment. These methods utilize intensity \citep{Masashi2005Building, Masashi2010Comparative, Matsuoka2010Estimation}, coherence \citep{Yonezawa2001Decorrelation, Arciniegas2007Coherence, Watanabe2016Detection, Wen2017Extraction}, and polarization features \citep{Yamaguchi2012Disaster, Chen2013Tsunami, Watanabe2016Detection, Karimzadeh2017Building} to assess building damage at a block unit level, depending on the acquisition mode. Several studies have attempted to extend the block-level approach and have explored new approaches at the building instance level using higher spatial resolution sensors such as COSMO-SkyMed and TerraSAR-X \citep{Liu2013Extraction, Brett2013Earthquake, Marco2015Identification, Ge2019Study}. DL-based methods have also been explored with SAR data to assess building damage \citep{Bai2018Framework, Adriano2019Multi, Bai2017Machine, Li2023DDFormer}. However, because of the lack of large-scale benchmark datasets, such as xBD in the optical domain, these methods have focused on local regions and single disaster events, and their ability to generalize to other disaster events remains largely unknown. 

\par The inherent challenges of SAR data, such as oblique viewing angles, speckle noise, object occlusion, and geometric distortions, complicate the accurate mapping of building damage compared to optical imagery \citep{ADRIANO2021132, xia2025openearthmap}. Furthermore, the limited availability of the VHR SAR data reduces its reliability as a source of pre-event data \citep{Brunner2010, ADRIANO2021132}. Considering these practical limitations, the most effective strategy for rapid assessment of building damage in all weather could arguably be to combine pre-event optical images, which provide accurate localization and detailed building information in the visible spectrum, with post-event SAR images, which capture structural information as a cue for building damage \citep{Adriano2019Multi}. Previous methods have attempted to align the two modalities with traditional statistical models \citep{Stramondo2006Satellite, Marco2009Exploiting, Brunner2010, Wang2012Postearthquake}. These statistical models are sensor-specific and require dedicated modeling for each sensor. DL methods offer a promising solution by automatically learning a high-dimensional feature space that aligns the two modalities. However, to train a DL model, one must have access to a high-quality, large-scale dataset with comprehensive coverage of various disaster events and sufficient geographic diversity. This remains a significant challenge that needs to be addressed.

\par To support AI-based research aimed at all-weather building damage mapping, we present \textsc{Bright}, the first open and globally distributed multimodal VHR dataset for building damage assessment. Advances in EO technology have enabled data providers like Capella\footnote{https://www.capellaspace.com/} and Umbra\footnote{https://umbra.space/} to offer VHR SAR imagery at a sub-meter level resolution per pixel. This allows for detailed building damage assessments at the individual building level, to guide targeted and effective rescue operations as required by emergency responders. Benefiting from the progress made in EO, \textsc{Bright} incorporates both pre-event optical imagery and post-event SAR imagery with spatial resolutions ranging from 0.3 meters to 1 meter per pixel. The types of disaster events considered in \textsc{Bright} are \textit{earthquakes}, \textit{storms (e.g., hurricane, cyclone)}, \textit{wildfires}, \textit{floods}, and \textit{volcanic eruptions}. These natural disasters accounted for 84$\%$ of the fatalities and 94$\%$ of the economic losses between 1998 and 2017 \citep{UNDRR2018EconomicLosses}. In addition to natural disasters, the \textsc{Bright} dataset further considers disasters caused by human activity, such as  \textit{accidental explosions} and  \textit{armed conflicts}, which also pose significant threats to human life and infrastructure and can occur unexpectedly, requiring a rapid response \citep{UNDRR2018Words, dietrich2025open}. The 14 disaster events cover 23 different regions distributed around the globe, with a focus on developing countries where external assistance is most urgently needed after a disaster. The labels are manually annotated with multi-level annotations that distinguish between damaged buildings and completely destroyed buildings. 

\begin{table*}[!t]
\scriptsize \renewcommand{\arraystretch}{1.4}
\caption{Comparison of \textsc{Bright} with the existing building damage assessment datasets. The OA indicates whether the dataset is open access (OA) or not, and GSD is an acronym for ground sampling distance (GSD). Note that since some datasets integrate other datasets, we summarize only the largest one to avoid duplication here. For example, the BDD dataset \citep{ADRIANO2021132} includes the Tohoku-Earthquake-2011 dataset \citep{Bai2018Framework} and Palu-Tsunami-2018 dataset \citep{Adriano2019Multi}.}\label{tbl:COMPARISON_WITH_OTHER_DATASETS}
  \centering
  
\begin{tabular}{l l l l l p{2.3cm} l l}
  \hline
 Dataset	& OA &  Modality & GSD (m/pixel) & No. of events  & Disaster type & No. of building & Granularity \\
 \hline
 \hline
ABCD \citep{Damage2017Fujita} & $\checkmark$ & Optical EO  & 0.4  & 1 & Tsunami  & N/A & Image-level \\ 
\citep{Nguyen2017Damage} & $\checkmark$ & Images on social media  & N/A  & 4 & 3 natural disasters  & N/A & Image-level \\
\citep{Cheng2021Deep} & $\checkmark$ & Optical EO   & N/A  & 1 & Hurricane  & 1,802 & Image-level \\ 
\citep{xue2024post} & $\checkmark$ & Street-view image  & N/A  & 1 & Hurricane  & 2,468 & Image-level \\ 
FloodNet \citep{Rahnemoonfar2021FloodNet} & $\checkmark$ & Optical EO  & N/A & 1 &  Flood  & 6,675 & Pixel-level\\
RescueNet \citep{Rahnemoonfar2023} & $\checkmark$ & Optical EO  & N/A & 1 &  Hurricane  & 10,903 & Pixel-level \\ 
Ida-BD \citep{Kaur2023Large} & $\times$ & Optical EO & 0.5 & 1 &  Hurricane  & 18,083 & Pixel-level \\ 
CRASAR-U-DROIDs  & \multirow{2}{*}{$\checkmark$} & \multirow{2}{*}{Optical EO} & \multirow{2}{*}{0.02-0.12} & \multirow{2}{*}{10} &  4 natural disasters  & \multirow{2}{*}{21,716} & \multirow{2}{*}{Pixel-level} \\ 
\citep{manzini2024crasar} &   &    &  &  & 1 man-made disaster &    &  \\
Noto-BDA-MV \citep{VescovoNoto2025} & $\checkmark$ &  Optical EO  &  N/A  & 1 & Earthquake & 140,208 & Pixel-level \\ 
xBD \citep{Gupta_2019_CVPR_Workshops} & $\checkmark$ &  Optical EO  &  <0.8  & 15 & 6 natural disasters & >700,000 & Pixel-level\\ 
QQB \citep{Sun2024Post} & $\checkmark$ & Optical and SAR EO & <1  & 1 & Earthquake  & 4,029 & Pixel-level\\ 
BDD \citep{ADRIANO2021132} & $\times$ & Optical and SAR EO & 1.2-3.3 & 9 & 3 natural disasters  & 123,453 & Pixel-level\\ 
 \hline
 \multirow{2}{*}{\textbf{\textsc{Bright}}} & \multirow{2}{*}{$\checkmark$} &  \multirow{2}{*}{Optical and SAR EO} & \multirow{2}{*}{0.3-1} & \multirow{2}{*}{14} & 5 natural disasters & \multirow{2}{*}{384,596} & \multirow{2}{*}{Pixel-level} \\ 
 &   &    &  &  & 2 man-made disasters &    &  \\
    \hline
  \end{tabular}
\end{table*}
\subsection{Comparison with existing datasets}
The comparison between \textsc{Bright} and existing datasets for building damage assessment is summarized in Table \ref{tbl:COMPARISON_WITH_OTHER_DATASETS}. Most current building damage assessment datasets are limited in scale and scope due to the limited availability of disaster events with corresponding open-source EO data and annotation efforts \citep{Rahnemoonfar2021FloodNet, Gupta2020RescueNet, Kaur2023Large}. Because of the high cost and time required for pixel-level labeling, some of the existing datasets provide image-level labeling, indicating only whether an image contains damaged buildings \citep{Damage2017Fujita, Nguyen2017Damage, Cheng2021Deep, xue2024post}. Although these image-level labeling datasets have served the community well, they lack the spatial precision needed to guide specific rescue operations. The xBD dataset \citep{Gupta_2019_CVPR_Workshops} is currently the largest open data collection, covering six natural disasters in 15 regions with more than 700,000 building instances. However, the xBD includes only optical EO data. It does not support all-weather disaster response. \cite{Sun2024Post} introduced a multimodal dataset, but it is limited to a single disaster event and contains only about 4,000 building instances. The small size makes it challenging to train DL models and limits the transferability of the trained models. 

\par The dataset most similar to \textsc{Bright} is the BDD proposed by \cite{ADRIANO2021132}. The main differences between BDD and \textsc{Bright} datasets are: (1) \textsc{Bright} covers more disaster events and building instances, including both natural and human-made disasters. (2) \textsc{Bright} has higher spatial resolution in both optical and SAR images. Whereas the highest resolution of SAR images in BDD is 1.2 meters, \textsc{Bright} provides finer detail with spatial resolutions ranging from 0.3 meters to 1 meter, enabling the detection of subtle structural damage in individual buildings. (3) Perhaps the most important difference is that whereas the re-distribution of BDD is restricted, \textsc{Bright} is an open-source dataset publicly available to the global community. Apart from the datasets listed in Table \ref{tbl:COMPARISON_WITH_OTHER_DATASETS}, there are other datasets targeted at monitoring hazardous events related to disasters, including landslides \citep{Landslide4Sense2022Ghorbanzadeh, Meena2023GLDD}, floods \citep{Bonafilia2020Sen1Floods11, Zhang2023Flood} and wildfires \citep{global2019Tomas, Huot2022Next, He2024global}, but are not related to building damage assessment. 

\begin{figure*}[!t]
    \centering
    \includegraphics[width=6.95in]{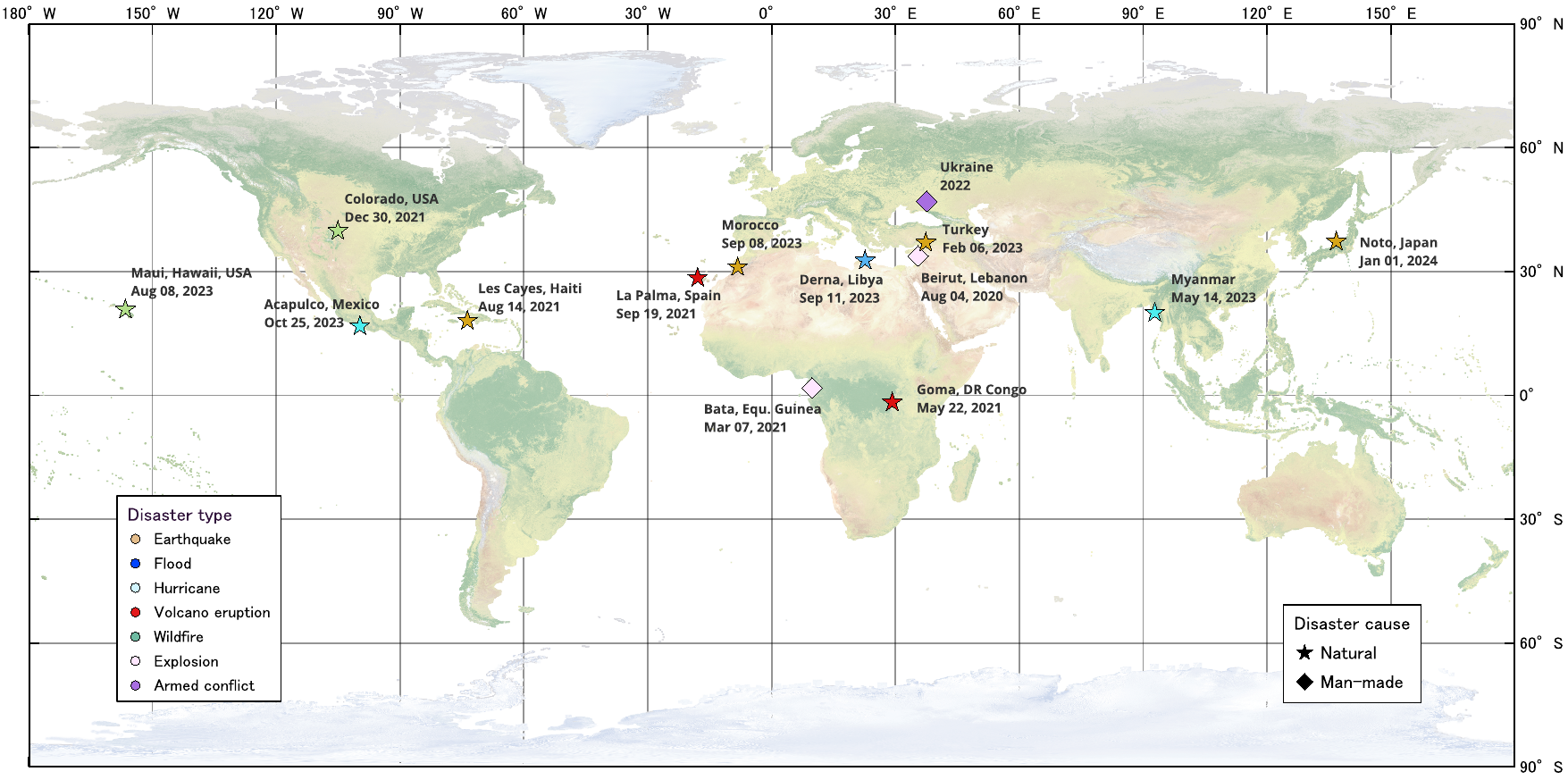}
   \caption{Geographic distribution of disaster events present in \textsc{Bright}. }
  \label{fig:events_geo_distribution}
\end{figure*}

\subsection{Main contribution}
The contributions of this paper are threefold:
\begin{enumerate}
    \item We present \textsc{Bright}, the first multimodal building damage dataset with sub-meter spatial resolution, which is publicly available to the community. \textsc{Bright} employs a combination of pre-event optical imagery and post-event SAR imagery, with various disaster events and rich geographic diversity, to support the study of AI-based multimodal building damage mapping, especially in developing countries.
    \item We evaluate a suite of contemporary models on \textsc{Bright} to establish robust baselines. Beyond supervised deep learning, \textsc{Bright} can support a wide range of AI-based methods. It enables research in unsupervised domain adaptation (UDA), semi-supervised learning (SSL), unsupervised multimodal change detection (UMCD), and unsupervised multimodal image matching (UMIM), among others. To demonstrate its utility, we benchmark a suite of representative models across several of these tasks. All experimental results, along with the source code and pretrained weights, are publicly released to provide strong baselines and accelerate future developments in disaster response in the community.
    \item We provide an in-depth analysis that uncovers key challenges and mechanisms of the multimodal building damage assessment task. Through carefully designed experiments, we reveal the difficulties of cross-event generalization, investigate the role of pre-event optical data in aiding damage classification, and quantify the performance gaps between different post-event modalities. These findings offer valuable insights for the development of more robust and practical models for disaster response.
    
\end{enumerate}

\begin{table*}[!t]
\scriptsize \renewcommand{\arraystretch}{1.75}
\caption{Summary of basic information of the \textsc{Bright} dataset with disaster events listed in chronological order. GSI refers to the Geospatial Information Authority of Japan, and IGN refers to the Instituto Geográfico Nacional (National Geographic Institute) of Spain.}\label{tbl:BRIGHT_info}
  \centering
  \begin{tabular}{l l l l l l l }
  \hline
 Disaster area	& Type of disaster & Date  &  GSD (m/pixel) &  Data provider / source   & No. of tiles  & No. of building \\
 \hline
\hline

 Beirut, Lebanon & Explosion (EP) &  04 Aug. 2020 & 1  & Maxar $\&$ Capella &   133 & 25,496 \\ 
\hline
 Bata, Equatorial Guinea  & Explosion (EP) & 07 Mar. 2021 & 0.5 & Maxar $\&$ Capella &    107 &  8,893 \\ 
\hline
Goma, DR Congo & Volcano eruption (VE) &  22 May 2021 &  0.33 & Maxar $\&$ Capella  &   123 & 18,741 \\ 
\hline
Les Cayes, Haiti & Earthquake (EQ) & 14 Aug. 2021 &  0.48 &  Maxar $\&$ Capella &   73 & 18,918 \\ 
\hline
La Palma, Spain & Volcano Eruption (VE)  &  19 Sept. 2021 - 13 Dec. 2021 &  0.3-0.35 & IGN (Spain) $\&$ Capella &  933 & 30,239 \\ 
\hline
Boulder, USA & Wildfire (WF)  & 30 Dec. 2021 - 01 Jan., 2022 &  0.6 & NAIP $\&$ Capella &  77 & 8,365 \\ 
\hline
Ukraine & Armed conflict (AC)  &  22 Mar. 2022 - 21 Sept. 2022  &  0.6  & Google Earth $\&$ Capella  & 513 & 56,770 \\ 
\hline
Turkey & Earthquake (EQ) & 06 Feb. 2023 & 0.30-0.35 &  Maxar $\&$ 
 Capella  $\&$  Umbra & 1,114 & 135,033  \\ 
\hline
Kyaukpyu, Myanmar & Cyclone (CC) & 14 May 2023 & 0.6  & Google Earth $\&$ Capella &  126 &  8,052 \\ 
\hline
Maui, Hawaii, USA & Wildfire (WF) &  08 Aug. 2023 - 09 Aug. 2023 &  0.6 & NOAA $\&$ Capella & 65 &  3,995 \\ 
\hline
Morocco & Earthquake (EQ) &  08 Sept. 2023 &  0.35-0.4 & Maxar $\&$ Capella &    567 & 6,269  \\ 
\hline
Derna, Libya & Flood (FL) &  10 Sept. 2023 & 0.35  &  Maxar $\&$ Capella & 124 & 1,0979 \\ 
\hline
Acapulco, Mexico & Hurricane (HC) & 25 Oct. 2023 &  0.35-0.8 & Google Earth $\&$ Capella & 212 & 18,437 \\    
\hline
Noto, Japan & Earthquake (EQ) & 01 Jan. 2024 &  0.5 & GSI $\&$ Umbra &  79 & 8,153 \\    

\hline
\hline
\emph{Total} & - & - & 0.3-1 &  -  &  4,246 & 384,596 \\      

    \hline
  \end{tabular}
\end{table*}


\begin{figure*}[!t]
    \centering
    \includegraphics[width=6.95in]{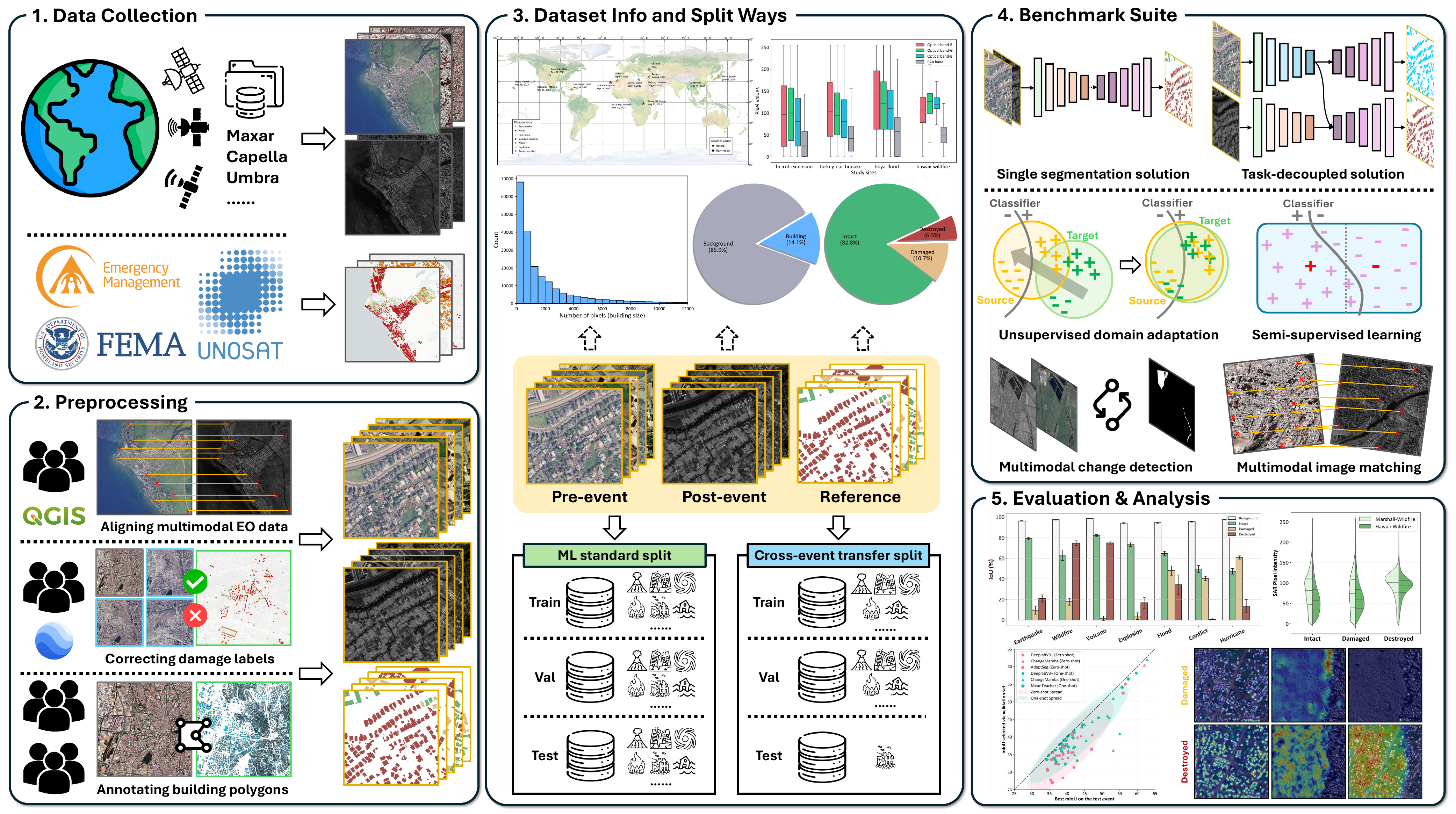}
   \caption{Overall flowchart of developing the \textsc{Bright} dataset.}
  \label{fig:bright_flowchart}
\end{figure*}

\begin{figure*}[!t]
    \centering
    \includegraphics[width=6.95in]{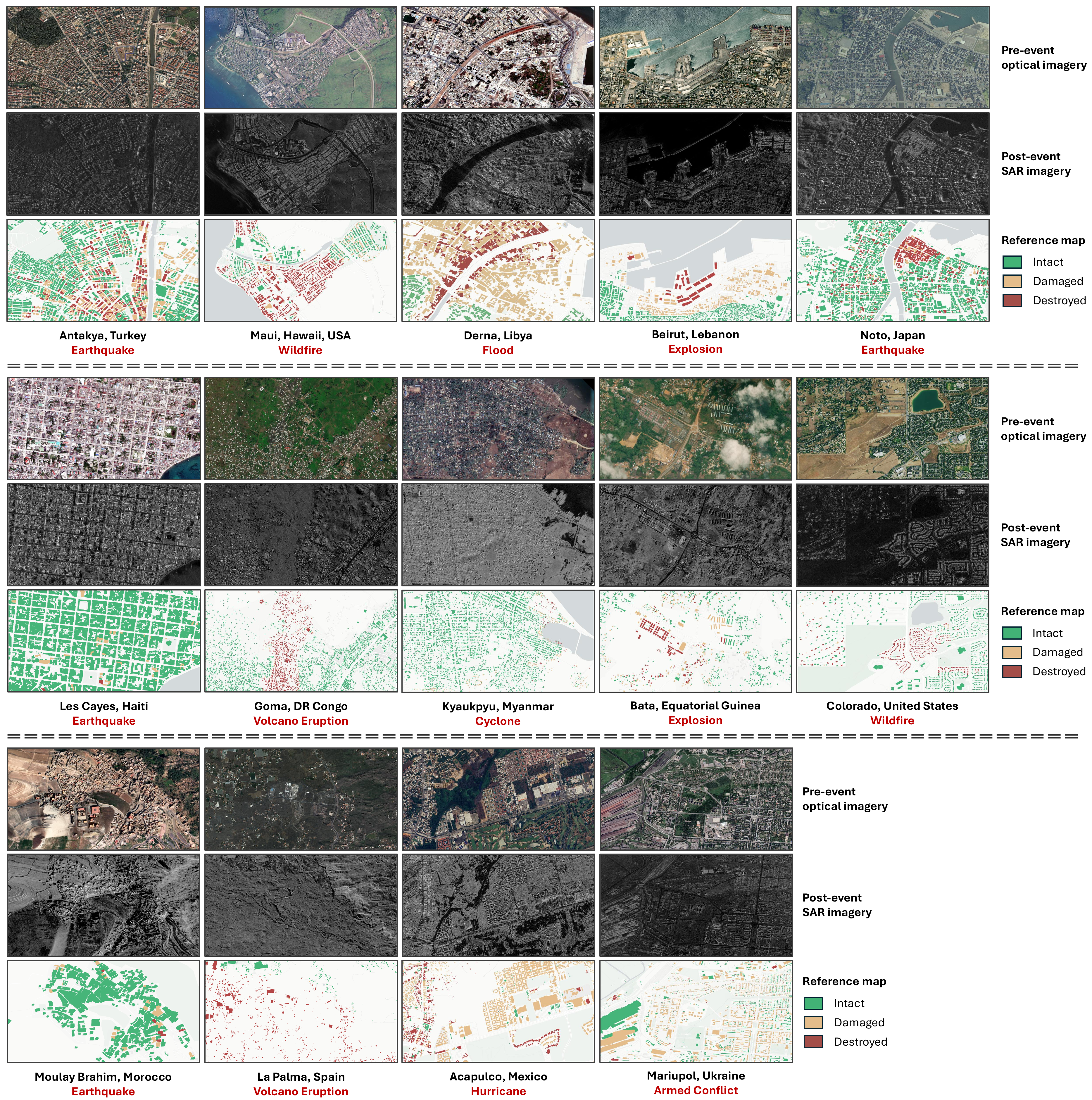}
   \caption{Thumbnails of local areas in 14 disaster events in the \textsc{Bright} dataset. The sources of EO images are illustrated in Table \ref{tbl:BRIGHT_info}. For visualization purposes, different events have different scales. }
  \label{fig:event_thumbnails}
\end{figure*}

\section{Dataset Description}
\subsection{Study areas and disaster events}
\par We selected 14 disaster events across the globe for \textsc{Bright}, as illustrated in Figure \ref{fig:events_geo_distribution} and Table \ref{tbl:BRIGHT_info}. Since both Capella Space and Umbra satellites were launched in 2020, we focused on study areas where disasters have occurred since then. The selected regions are primarily in developing countries, where public administration and disaster response capacities tend to be weaker compared to those in developed nations, making international assistance more critical. The dataset covers five major types of natural disasters: earthquakes, storms (including hurricanes and cyclones), wildfires, floods, and volcanic eruptions. Additionally, it includes human-made disasters, such as accidental explosions and armed conflicts. Detailed descriptions of the 14 disaster events are provided in Appendix \ref{app:disaster_events_details}.

\subsection{Construction of \textsc{Bright}}
Figure \ref{fig:bright_flowchart} shows the flowchart of developing \textsc{Bright}. The optical EO data in the dataset are mainly from Maxar's Open Data program\footnote{\url{https://www.maxar.com/open-data}}, while the SAR EO data are from Capella Space\footnote{\url{https://www.capellaspace.com/earth-observation/gallery}} and Umbra\footnote{\url{https://umbra.space/open-data/}}. Both Capella and Umbra data have two imaging modalities, \emph{i.e.}, Spotlight and Stripmap, respectively. The Spotlight mode has a higher spatial resolution but less coverage. In the region of interest, we preferred Spotlight mode if suitable data was available in the data provider's inventory. Otherwise, we chose Stripmap. The optical EO data consists of red, blue, and green bands, while the SAR EO data consists of amplitude data in the VV or HH bands. For optical EO data, the digital number was converted to reflectance and then standardized to an 8-bit data format. For SAR imagery, after the data had been terrain-corrected, we utilized the pre-processed 8-bit data when available. In cases where 8-bit data was not provided, we employed the data provider's recommended method\footnote{https://support.capellaspace.com/scaling-geo-images-in-qgis} to convert the amplitude data. Although both optical and SAR images are geocoded, there are still pixel offsets between them. Therefore, multiple EO experts manually aligned the paired optical and SAR data and cross-checked their results to ensure the precise registration between the two modalities. Figure \ref{fig:feature_points_for_UMIM} in Appendix \ref{app:registration_error_estimate} shows the selected control points on three disaster scenes.

\begin{table}[!t]
  \renewcommand{\arraystretch}{1.25}
  \caption{Definition of different categories of damage in \textsc{Bright}.}
  \label{tbl:damage_classes_definition}
  \centering
  \begin{tabular}{l | p{5.5cm} }
    \hline	
  Category &  Definition  \\
    \hline
    \hline
   Background (0) & All non-building pixels \\
     \hline
    \rowcolor{green!13} Intact (1) & No visible signs of structural damage, water intrusion, shingle displacement, or burn marks.   \\
    \hline
    \rowcolor{orange!13} Damaged (2) & Partial structural damage to the building, such as missing roof members, visible cracks, or partial wall/roof collapse. Buildings may be partially burned, surrounded by water or mud, or affected by nearby volcanic flows.  \\
    \hline
    \rowcolor{red!13} Destroyed (3) & Completely collapsed, burned, partially/completely covered by water/mud or no longer present  \\
   
    \hline
\end{tabular}
\end{table}

\par The labels in \textsc{Bright} consist of two components: building polygons and post-disaster building damage attributes. Expert annotators manually labeled the building polygons, then all labels underwent independent visual inspections of EO experts to ensure accuracy. Damage annotations were obtained from Copernicus Emergency Management Service\footnote{\url{https://emergency.copernicus.eu}}, the United Nations Satellite Centre (UNOSAT) Emergency Mapping Products\footnote{\url{https://unosat.org/products}}, and the Federal Emergency Management Agency (FEMA)\footnote{\url{https://www.fema.gov}}. These annotations were derived through visual interpretation of high-resolution optical imagery captured before and after the disasters by EO experts, supplemented by partial field visits. To harmonize these diverse annotations and ensure consistency across all 14 disaster events, we implemented a rigorous, multi-stage process. First, we established a single, standardized three-tier classification scheme, including Intact (with pixel value 1), Damaged (with pixel value 2), and Destroyed (with pixel value 3), with clear definitions provided in Table \ref{tbl:damage_classes_definition}, drawing on the frameworks of FEMA’s Damage Assessment Operations Manual, EMS-98, the BDD dataset \citep{ADRIANO2021132}, and the xBD dataset \citep{Gupta_2019_CVPR_Workshops}. While the source agencies' terminology can differ (e.g., ``Severe Damage'' vs. ``Major Damage''), their underlying definitions for EO-based assessment are conceptually consistent. We leveraged this alignment for an initial rule-based mapping, where various intermediate damage tiers were conservatively aggregated into our single ``Damaged'' category. Second, our team of EO experts conducted a comprehensive manual verification and refinement of every annotation using multi-temporal VHR imagery on platforms like Google Earth Pro. This final stage served as the ultimate guarantor of consistency. We paid special attention to ambiguous source labels, such as ``Possibly Damaged''. Adopting a conservative approach, these were re-classified as ``Intact'' if clear structural damage was not evident, thereby ensuring a high-confidence ``Damaged'' class. We also manually disaggregated all area-based annotations (i.e., where an entire block was assigned a single category). We re-processed these to assign a precise, building-wise damage label to each individual structure, ensuring instance-level consistency and granularity across the entire dataset. The damage annotations were provided as vector point files. The final building damage labels were generated by overlaying these points withd the building polygons and assigning corresponding damage attributes. To prevent geographic misallocation due to possible coordinate offsets, the coordinate systems of the points and polygons were unified, with a visual inspection performed prior to the final allocation. Figure \ref{fig:event_thumbnails} presents thumbnails of selected local areas from the 14 disaster events.

\subsection{Statistics of \textsc{Bright}}
\begin{figure*}[!t]
    \centering
    \includegraphics[width=6.9in]{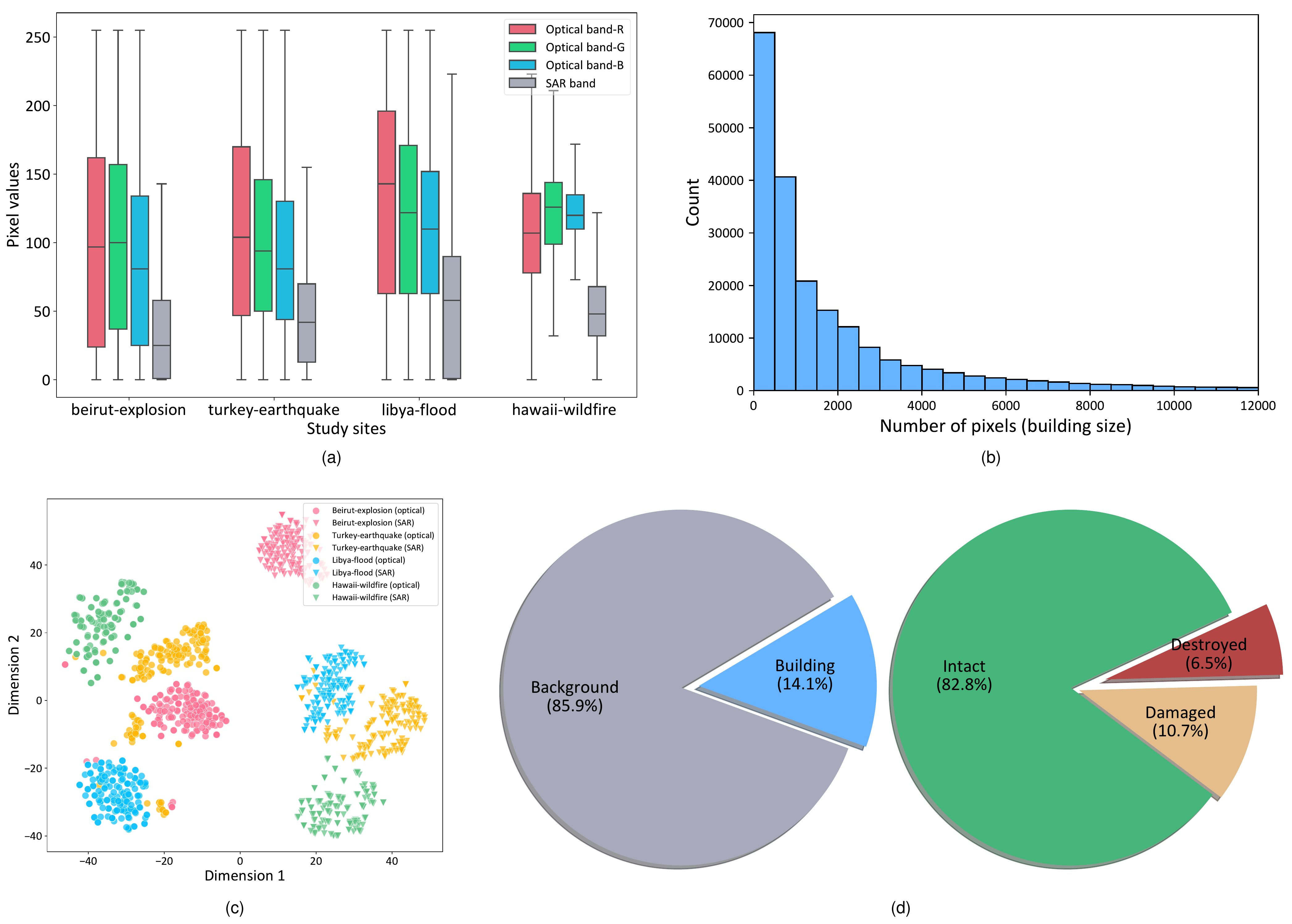}
   \caption{Statistics of the \textsc{Bright} dataset. (a) Distributions of band values of samples from four study sites. (b) Distribution of building scales. (c) Feature distribution of buildings of four events under two imaging modalities. (d) Percentage of building and background pixels and percentage of different damage levels in building pixels.}
  \label{fig:BRIGHT_stat_info}
\end{figure*}

\par The basic information about \textsc{Bright}, including disaster events, EO data, the number of corresponding EO tiles, and the total number of building pixels, is summarized in Table \ref{tbl:BRIGHT_info}. After cropping the EO data into 1024$\times$1024-pixel tiles, \textsc{Bright} contains 4,246 multimodal image pairs. 


\par The key statistics of \textsc{Bright} are illustrated in Figure \ref{fig:BRIGHT_stat_info}. Figure \ref{fig:BRIGHT_stat_info}-(a) shows the pixel value distribution for optical and SAR images from one human-made disaster and three natural disasters. The varying geographical landscapes and land cover across different regions result in distinct means and standard deviations of pixel values. This highlights \textsc{Bright}’s geographical diversity, which makes it a robust dataset for studying building damage assessment in diverse environments. To ensure that models trained on \textsc{Bright} can accurately detect buildings and assess damage levels, it is crucial that the dataset includes a wide variety of building styles from different regions. Figure \ref{fig:BRIGHT_stat_info}-(b) shows that \textsc{Bright} covers buildings at multiple scales, exhibiting a “long-tail” distribution. This multi-scale representation challenges DL models to develop the ability to capture features at varying scales, enhancing robustness and accuracy. 

\par Figure \ref{fig:BRIGHT_stat_info}-(c) further illustrates the feature distribution of buildings in the optical and SAR images for the four events shown in Figure \ref{fig:BRIGHT_stat_info}-(a), which demonstrates clear inter-event separability in both modalities. \textsc{Bright} also faces a significant challenge of sample imbalance, as shown in Figure \ref{fig:BRIGHT_stat_info}-(d). There is a notable imbalance between background pixels and foreground (building) pixels, with a ratio of approximately 7:1. The imbalance exists within the damage categories: about 6.5$\%$ of building pixels represent destroyed buildings, 10.7$\%$ correspond to damaged buildings, and 82.8$\%$ are intact buildings. This imbalance can complicate model training, necessitating careful strategies to develop robust DL models.

\begin{table}[!t]
  \renewcommand{\arraystretch}{1.3}
  \caption{Proxy registration errors (in pixel) estimated using different multimodal image descriptors. The estimated registration errors for each event are reported in Table \ref{tbl:registration_error_each_event} in Appendix \ref{app:registration_error_estimate}.}
  \label{tbl:registration_error}
  \centering
  \begin{tabular}{l | c c}
    \hline	
  Descriptor &   RMSE (All) &   RMSE (Building)  \\
    \hline
    \hline
    RIFT \citep{Li2020RIFT} & 1.125 & 1.138 \\
    LNIFT \citep{Li2022LNIFT} & 0.857 & 0.825 \\
    SRIF \citep{Li2023Multimodal} & 1.090 & 1.054  \\
    \hline
    Average &  1.024  & 1.006 \\
    \hline
\end{tabular}
\end{table}

\par Moreover, since accurate registration ensures spatial consistency across modalities, the registration accuracy between optical and SAR EO data in \textsc{Bright} was analyzed to provide a solid foundation for accurate building damage assessment. Due to the absence of real ground truth, a proxy method that leverages existing multimodal image descriptors was introduced to estimate registration errors. This approach is detailed in Appendix \ref{app:registration_error_estimate}. The Table \ref{tbl:registration_error} reports the mean registration errors, measured as the root mean square error (RMSE) of pixel displacements. It was obtained using three representative multimodal image registration methods: RIFT~\citep{Li2020RIFT}, SRIF~\citep{Li2023Multimodal}, and LNIFT~\citep{Li2022LNIFT}. The overall average RMSE is approximately 1.024 pixels, with a lower error of 1.006 pixels specifically within the building regions.

\subsection{Dataset splitting strategy}\label{sec:dataset_split}
To train DL models using \textsc{Bright} and evaluate their generalizability, it is necessary to split the dataset into a training set, validation set, and test set. \cite{gerard2024simple} suggested that dividing the dataset on an event-by-event basis, rather than randomly across the entire dataset, provides a more accurate reflection of a model’s generalizability. Therefore, for the 14 events listed in Table \ref{tbl:BRIGHT_info}, we divide the corresponding data for each event into a ratio of 7:1:2 for training, validation, and test subsets, respectively. Then, the subsets obtained for each event are merged to create the final training, validation, and test sets. In the experiments, the baseline models are trained using the training set, and the optimal hyperparameters (e.g., learning rate) and checkpoints are selected based on performance on the validation set. The generalization capability of the baseline models is subsequently evaluated on the test set. 

\par In addition to the above standard ML data splitting, we also introduce a cross-event transfer setup to better evaluate the ability of models to generalize across disaster events. This is a critical challenge in real-world applications where models are expected to handle unseen disaster types and locations. Two setups are established for cross-event transfer generalization:
\begin{list}{\textbullet}{}
    \item Zero-shot setup: This setting mimics a real-world scenario where a newly occurring disaster must be analyzed without any prior labeled data from the same event. We isolate one event as an unseen test set while using the remaining 13 events for training and validation. This setting evaluates the cross-event generalization ability of models, testing how well learned knowledge can be transferred from previous disasters to an entirely new disaster event. Due to the high variability of disaster types and geographies, this setup is inherently challenging, as models trained on past disasters may struggle to assess damage patterns in a previously unseen event accurately.
    \item One-shot setup: Recognizing the difficulty of the zero-shot setup, we introduce a one-shot setup. This setting simulates a realistic scenario where a single, representative sample from the new disaster can be quickly labeled to guide model adaptation. In this setting, a limited subset of labeled data (one pair for training and one pair for validation) from the target disaster event is incorporated into the training process. At the same time, the majority of the test set remains unseen. This setup evaluates the model’s ability to leverage a minimal amount of manually labeled data to improve disaster-specific adaptation. 
\end{list} 

\par It is worth noting that our cross-event transfer setup differs from classic few-shot learning tasks in the computer vision field \citep{One2017Amirreza, Wang2020Generalizing}. Our goal is not to recognize new classes, but to adapt the model's knowledge of existing classes to a new domain, \emph{i.e.}, an unseen disaster event.

\section{Methodology}
\subsection{Problem statement}\label{sec:4.1}
The objective of building damage assessment is to interpret EO data of areas affected by a disaster by generating a building damage map that reflects the extent of damage to buildings. To achieve this, two common approaches are typically employed. One is to directly treat the building damage assessment task as a single semantic segmentation task \citep{ADRIANO2021132, Gupta2020RescueNet}. In this approach, the pre- and post-event images are taken as inputs of the model, and then the final damage map is directly predicted. This process can be formalized as $\mathbf{Y}^{dam} = \mathcal{M}^{seg}(\mathbf{X}^{\mathrm{T}_{1}}, \mathbf{X}^{\mathrm{T}_{2}})$, where $\mathbf{X}^{\mathrm{T}_{1}}$ is the pre-event imagery, $\mathbf{X}^{\mathrm{T}_{2}}$ is the post-event imagery, $\mathcal{M}^{seg}\left(\cdot\right)$ is a semantic segmentation model, $\mathbf{Y}^{dam}$ is the obtained damage map. In the context of this paper, $\mathbf{X}^{\mathrm{T}_{1}}$ is VHR optical imagery and $\mathbf{X}^{\mathrm{T}_{2}}$ is VHR SAR imagery. 

\par The second adopts the task decoupling approach \citep{Gupta_2019_CVPR_Workshops, ZHENG2021Building}, which breaks down building damage assessment into two subtasks: the building localization task, \emph{i.e.}, separating the building from the background, and the damage classification task, \emph{i.e.}, focusing on the classification between different levels of damage. This approach can be formulated as $\mathbf{Y}^{loc} = \mathcal{M}^{loc}(\mathbf{X}^{\mathrm{T}_{1}})$ and $\mathbf{Y}^{clf} = \mathcal{M}^{clf}(\mathbf{X}^{\mathrm{T}_{1}}, \mathbf{X}^{\mathrm{T}_{2}})$, where $\mathbf{Y}^{loc}$ is the building localization map, $\mathbf{Y}^{clf}$ is the damage classification map, $\mathcal{M}^{loc}\left(\cdot\right)$ and $\mathcal{M}^{clf}\left(\cdot\right)$ are models for building localization and damage classification, respectively. $\mathcal{M}^{loc}\left(\cdot\right)$ and $\mathcal{M}^{clf}\left(\cdot\right)$ can be two separate models \citep{Gupta_2019_CVPR_Workshops} or a unified multi-task learning model \citep{ZHENG2021Building, Chen2022Dual, Chen2024ChangeMamba}. The final building damage map is obtained by combining the two outputs using a simple mask operation: $\mathbf{Y}^{dam}=\mathbf{Y}^{loc} \odot \mathbf{Y}^{clf}$. Since this work aims not only to provide a large-scale multimodal dataset to support all-weather disaster response, but also to offer insights for designing appropriate methods in future research, both approaches are employed in the experiments to compare their results.

\par It is worth noting that in this work, we focus on the formulation of building damage assessment as a bi-temporal task, where both pre- and post-event images are used as inputs. This formulation aligns closely with generic change detection tasks, which aim to identify changes between two time points. Conceptually, building damage assessment can be viewed as a specialized ``one-to-many'' semantic change detection problem \citep{ZHENG2021Building, Zheng2024Unifying, Lu2024Bitemporal}, where the objective is not only to detect whether a change has occurred but also to categorize the type and severity of changes (damages) to buildings. Many existing methods are thus derived from or adapted versions of generic change detection frameworks \citep{Chen2024ChangeMamba, Zheng2024Unifying, Guo2024SAAN}.

\subsection{Benchmark suites}
Several advanced deep network architectures from both the computer vision and EO communities are evaluated on \textsc{Bright}. Since building damage assessment can be considered a specialized semantic segmentation task, we adopted two well-known segmentation networks from the computer vision field: UNet \citep{Ronneberger2015UNet} and DeepLabV3+ \citep{Chen2018Encoder}; and five state-of-the-art networks from the EO community: SiamAttnUNet \citep{ADRIANO2021132}, SiamCRNN \citep{Chen2019a}, ChangeOS \citep{ZHENG2021Building}, DamageFormer \citep{Chen2022Dual}, and ChangeMamba \citep{Chen2024ChangeMamba}. These seven networks encompass a broad range of representative DL architectures, including convolutional neural networks (CNNs), recurrent neural networks (RNNs), Transformers, and the more recent Mamba architecture. Among the seven networks, UNet, DeepLabV3+, and SiamAttnUNet adopt the first approach defined in Section~\ref{sec:4.1}, \emph{i.e.}, directly treating building damage assessment as a single semantic segmentation task. In contrast, SiamCRNN, ChangeOS, DamageFormer, and ChangeMamba adopt the second approach by decoupling the task into building localization and damage classification tasks.

\par Beyond supervised DL models, \textsc{Bright} also enables the evaluation of other learning strategies and methods commonly explored in the EO and computer vision communities:
\begin{list}{\textbullet}{}
    \item Unsupervised domain adaptation (UDA) methods for the zero-shot transfer setup, enabling models to transfer knowledge across disaster events with no labeled samples from the target event.
    \item Semi-supervised learning (SSL) approaches for the one-shot transfer setup, leveraging a small number of labeled samples and the remaining unlabeled samples from new disaster events to refine model adaptation.
    \item Unsupervised multimodal change detection (UMCD) methods, which exploit the modality-independent relationship in optical and SAR data to detect land-cover changes without requiring manual annotations.
    \item Unsupervised multimodal image matching (UMIM) methods, which aim to learn modality-independent features to enable automatic registration of multimodal data without relying on manual alignment. 
\end{list}

\begin{table*}[!t]
  \renewcommand{\arraystretch}{1.25}
  \caption{Accuracy assessment for different DL models on the test set under the standard ML data split (set-level mIoU). The highest values are highlighted in \textbf{\textcolor{purple}{purple}}, and the second-highest results are highlighted in \textbf{\textcolor{teal}{teal}}.}
  \label{tbl:bda_BRIGHT_all}
  \centering
  \begin{tabular}{p{2.9cm} | c c c c c c c c}
    \hline	
   \multirow{2}{*}{Method}  & \multirow{2}{*}{$F_{1}^{loc}$ ($\%$)}   & \multirow{2}{*}{$F_{1}^{clf}$ ($\%$)}    & \multirow{2}{*}{Final OA ($\%$)} & \multirow{2}{*}{Final mIoU ($\%$)}  & \multicolumn{4}{c}{IoU per class ($\%$)}  \\
    \cline{6-9} 
    & &  & & & Background  &  Intact &  Damaged & Destroyed  \\
    \hline
    \hline
      UNet  &	87.97 &	72.24 & 95.47	& 64.94 & 96.19	& 71.27	& 39.13 & 53.17 \\ 
      DeepLabV3+  & 87.00 &	70.33 & 95.43 & 64.80 &	95.98 &	70.98 &	37.53 & 54.69 \\ 
      SiamAttnUNet  &  88.16	&	70.13 &	95.45 &	64.26 &	96.14 &	71.90 &	35.06 & 53.92 \\ 
      \hline
      SiamCRNN  & 89.45 &	72.02 &	95.76 &	65.73 & 96.48 & 73.44 &	38.91 & 54.18 \\ 
      ChangeOS  &	89.60 &	71.88 &	95.84 &	65.98 &	96.54 &	73.85 &	38.99 & 54.53 \\ 
      DamageFormer & \textcolor{teal}{\textbf{90.29}} &	\textcolor{teal}{\textbf{72.51}} &	\textcolor{teal}{\textbf{96.13}} &	\textcolor{teal}{\textbf{67.09}} & \textcolor{teal}{\textbf{96.87}} & \textcolor{teal}{\textbf{75.04}}	& \textcolor{teal}{\textbf{39.86}}	& \textcolor{teal}{\textbf{56.59}} \\ 
      ChangeMamba & \textcolor{purple}{\textbf{90.90}} & \textcolor{purple}{\textbf{72.70}} & \textcolor{purple}{\textbf{96.22}} & \textcolor{purple}{\textbf{67.63}} & \textcolor{purple}{\textbf{96.96}} & \textcolor{purple}{\textbf{75.59}} & \textcolor{purple}{\textbf{40.05}} & \textcolor{purple}{\textbf{57.91}} \\ 

    \hline
\end{tabular}
\end{table*}

\subsection{Model training}
To train the supervised models, we use a combination of cross-entropy loss and Lovasz softmax loss \citep{Berman_2018_CVPR}. Cross-entropy loss serves as the basic loss function for dense prediction tasks, while Lovasz softmax loss effectively addresses sample imbalance between non-building and building pixels and across different damage levels. For UNet, DeepLabV3+, and SiamAttnUNet, which directly predict damage maps from the input multimodal image pairs, the training loss function is defined as:
\begin{equation} 
    \mathcal{L}^{bda}_{coupled} = \mathcal{L}^{bda}_{ce} + \mathcal{L}^{bda}_{lov}. 
\end{equation}
For SiamCRNN, ChangeOS, DamageFormer, and ChangeMamba, which decouple building damage assessment into building localization and damage classification subtasks, the training loss function is defined as:
\begin{equation}
    \mathcal{L}^{bda}_{decoupled} = \mathcal{L}^{loc}_{ce} + \mathcal{L}^{loc}_{lov} + \mathcal{L}^{clf}_{ce} + \mathcal{L}^{clf}_{lov}. 
\end{equation}

\par All the models are trained using the AdamW optimizer \citep{loshchilov2017decoupled} with a learning rate of 1$e^{-4}$ and a weight decay of 5$e^{-3}$. The training process consists of 50,000 iterations, with a batch size of 16. To enhance sample diversity and improve model generalization, we apply several data augmentation techniques, including random flipping, random rotation (in 90-degree increments), and random cropping. For the zero-shot and one-shot setups, we also test UDA and SSL techniques to better utilize unlabeled and limited target disaster data. The training settings for the UDA, SSL, UMCD, and UMIM approaches are provided in Appendix \ref{app:UDA}, \ref{app:SSL}, \ref{app:UMCD}, and \ref{app:UMIM}, respectively.

\subsection{Accuracy assessment}
\par We adopt overall accuracy (OA), F1 score (F1), and mean intersection over union (mIoU) to evaluate the performance of the models. These are commonly used metrics in building damage assessment \citep{ZHENG2021Building}. Following the setup in previous unimodal building damage assessment studies and the related xView2 Challenge \citep{Gupta_2019_CVPR_Workshops}, the F1 score is used to assess the performance of the models in the building localization and damage classification subtasks. OA and mIoU are used to measure the overall quality of the building damage map, providing a comprehensive assessment of the models' ability to localize buildings and classify damage levels accurately.

\section{Results and Analysis}

\subsection{Evaluation on standard machine learning data split}

Table \ref{tbl:bda_BRIGHT_all} shows the results for each model on the test set. We observe that ChangeMamba achieves the best overall performance, with an OA of 96.22$\%$, a mIoU of 67.63$\%$, and the highest $F_{1}^{loc}$ and $F_{1}^{clf} $scores of 90.90$\%$ and 72.70$\%$, respectively. DamageFormer also performs well, following ChangeMamba, with a mIoU of 67.09$\%$ and an OA of 96.13$\%$. Both models demonstrate a strong capability in the building localization and damage classification tasks. The accuracy of ChangeMamba and DamageFormer underscores the importance of leveraging advanced DL architectures to improve performance in complex tasks such as building damage assessment. For models that use a direct prediction approach (UNet, DeepLabV3+, SiamAttnUNet), UNet achieves the best results, with a mIoU of 64.94$\%$ and an OA of 95.47$\%$. However, its performance still lags behind the decoupled models, which emphasizes the advantage of task decoupling.

\begin{table*}[ht]
  \renewcommand{\arraystretch}{1.25}
  \caption{The mIoU on different events for different DL models (event-level mIoU). The highest values are highlighted in \textbf{\textcolor{purple}{purple}}, and the second-highest results are highlighted in \textbf{\textcolor{teal}{teal}}.}
  \label{tbl:bda_BRIGHT_event}
  \centering
  \begin{tabular}{l | c c c | c c c c}
    \hline	
   Events & UNet  & DeepLabV3+  & SiamAttnUNet  & SiamCRNN   & ChangeOS   & DamageFormer  & ChangeMamba  \\
    \hline
    \hline
     Beirut-EP-2020  & 52.23	& 48.53 & 51.85 & 52.04 &  \textcolor{teal}{\textbf{52.57}} & \textcolor{purple}{\textbf{60.55}} & 49.38  \\ 
     Bata-EP-2021   & 39.24 & 40.26 & 38.69 & \textcolor{purple}{\textbf{41.44}} & 39.80 & 40.37 &  \textcolor{teal}{\textbf{40.47}} \\ 
     Goma-VE-2021  &	61.52 & 59.26 & \textcolor{purple}{\textbf{62.95}} & 62.50 & 61.84 & 61.55 &  \textcolor{teal}{\textbf{62.89}} \\ 
     Les Cayes-EQ-2021   & 40.39 & 38.52 &40.84 & 40.99 & 41.24  & \textcolor{teal}{\textbf{41.75}} & \textcolor{purple}{\textbf{42.36}}  \\ 
     La Palma-VE-2021  &	62.72 & 64.31 & \textcolor{purple}{\textbf{65.56}} & 64.24 & \textcolor{teal}{\textbf{65.34}} & 64.27 &  65.13 \\ 
     Marshall-WF-2021   & 53.96 &  44.67 & 51.48 & 56.57 & 56.55  & \textcolor{purple}{\textbf{57.50}} & \textcolor{teal}{\textbf{57.23}} 	 \\ 
     Ukraine-AC-2022   &	43.92 & 43.84 & 40.71 & 45.52 & 45.56 & \textcolor{purple}{\textbf{47.45}} &   \textcolor{teal}{\textbf{47.43}} \\ 
     Turkey-EQ-2023   & 49.34 & 49.53 & 48.57 & 50.69 & \textcolor{teal}{\textbf{52.34}} & 51.36 &  \textcolor{purple}{\textbf{52.70}}	 \\ 
     Kyaukpyu-CC-2023   &	38.11 & 38.56 & 40.49 & \textcolor{teal}{\textbf{42.35}} &  42.03 & \textcolor{purple}{\textbf{48.42}} & 41.05  \\ 
     Hawaii-WF-2023  & 53.02 & 57.81 & 57.92 & 55.34 &  58.40 & \textcolor{purple}{\textbf{60.82}} & \textcolor{teal}{\textbf{60.70}} \\ 
     Morocco-EQ-2023   &	42.62 & 42.05 & 42.58 & 43.17 &  43.05 & \textcolor{teal}{\textbf{44.00}} &  \textcolor{purple}{\textbf{44.06}} \\ 
     Derna-FL-2023  & 59.47 &  56.08 & \textcolor{purple}{\textbf{66.82}} & 59.48 & 57.73 & 62.24 &  \textcolor{teal}{\textbf{63.30}}  \\ 
     Acapulco-HC-2023 & 49.21 & 46.66 & 42.15 & 49.98 &  \textcolor{purple}{\textbf{51.08}} & \textcolor{teal}{\textbf{50.82}} & 47.01 \\ 
     Noto-EQ-2024  &	36.49 & 37.33 & 36.56 & 38.35 & \textcolor{teal}{\textbf{42.84}} & 40.57 & \textcolor{purple}{\textbf{45.78}} \\ 
    \hline
    Average  &	48.73 & 47.67 & 49.08 & 50.19 & 50.74& \textcolor{purple}{\textbf{52.26}} & \textcolor{teal}{\textbf{51.39}}  \\
    \hline
\end{tabular}
\end{table*}
\par
\begin{figure*}[ht]
    \centering
    \includegraphics[width=6.1in]{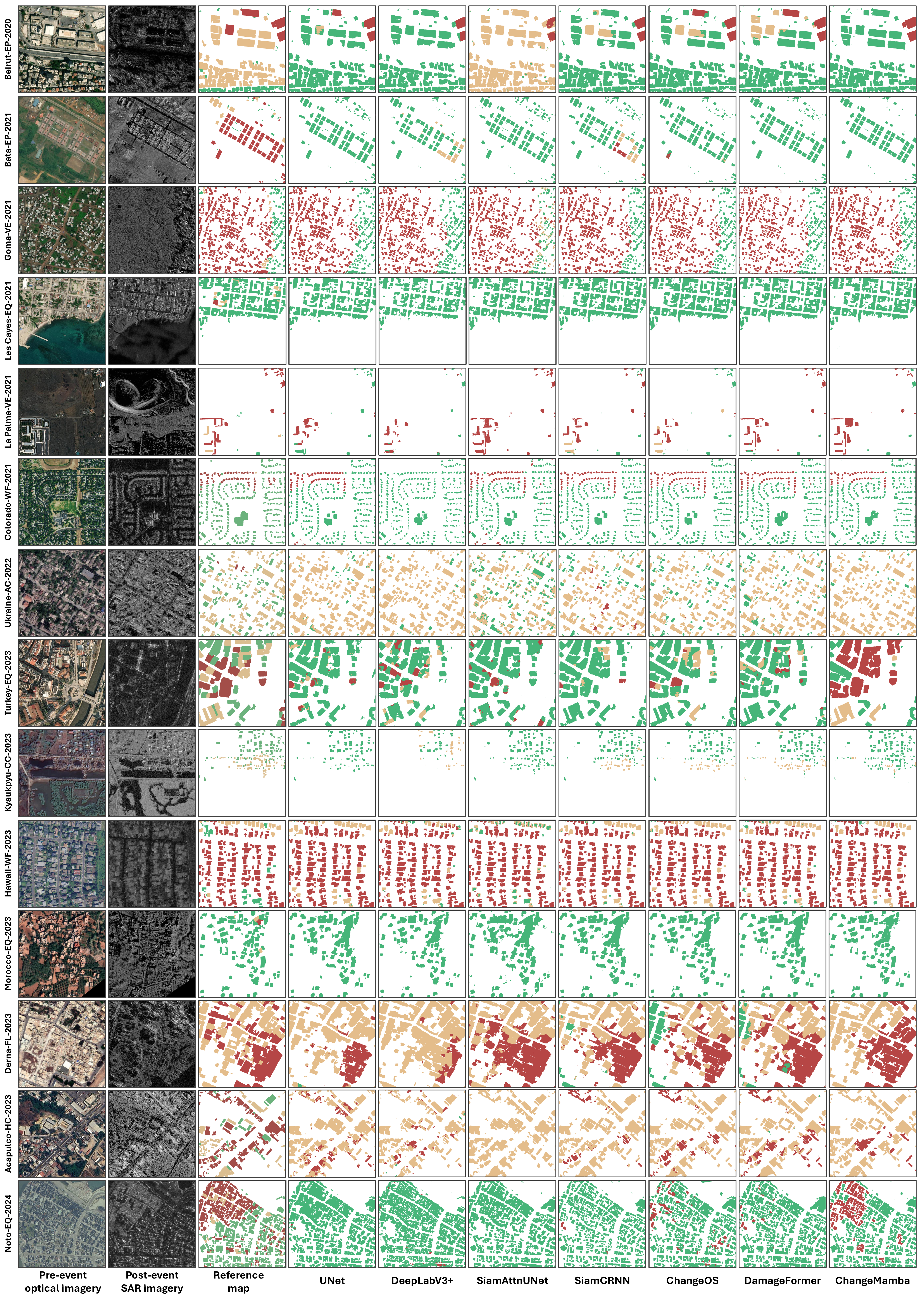}
   \caption{Damage maps predicted by different models on the test images of the 14 disaster events. The meaning of the color in reference maps and damage maps is consistent with Figure \ref{fig:event_thumbnails}. The sources of EO images are illustrated in Table \ref{tbl:BRIGHT_info}. }
  \label{fig:prediction}
\end{figure*}

 To ensure that the evaluation is not dominated by a few events with a large number of images, e.g., Turkey-EQ-2023, Table \ref{tbl:bda_BRIGHT_event} presents the event-level mIoU for each model. ChangeMamba and DamageFormer achieve the highest average mIoU, with scores of 51.39$\%$ and 52.26$\%$, respectively. DamageFormer performs very well on events such as Beirut-EP-2020, Marshall-WF-2021, and Derna-FL-2023. This shows its robustness across different types of disasters. Although performance varies across events, earthquake-related events such as Les Cayes-EQ-2021, Morocco-EQ-2023, and Noto-EQ-2024 present a greater challenge to all models, with a relatively low average mIoU. This highlights the need for further research to improve the robustness of earthquake damage assessment models, particularly where damage patterns are more complex and diverse. Figure \ref{fig:prediction} shows some building damage maps obtained by the seven models on the test set. 

\subsection{What have the models learned and what can they learn?}

\begin{figure*}[!t]
   \centering
    \includegraphics[width=6.85in]{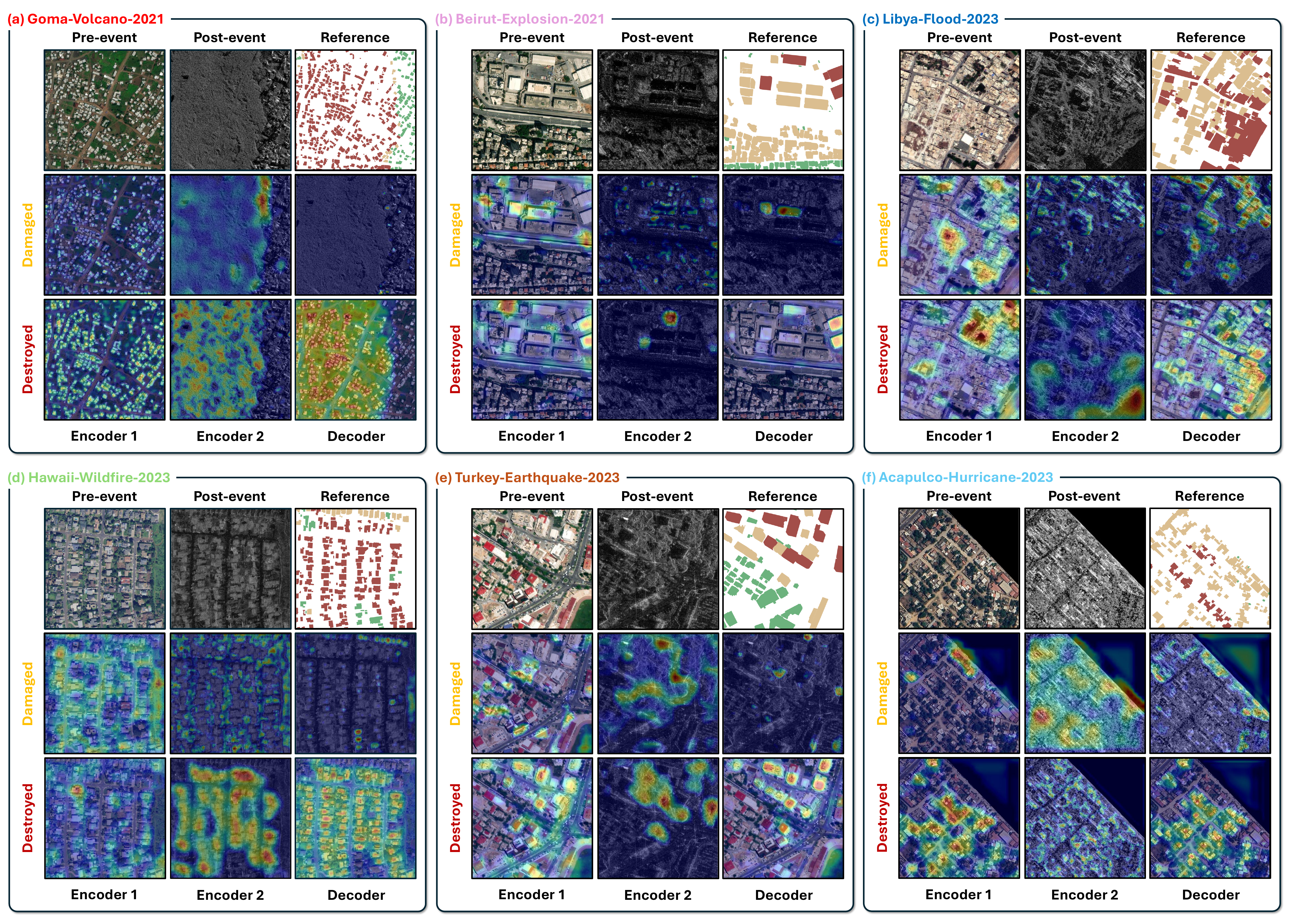}
   \caption{Visualization of feature response to “Damaged” and “Destroyed” categories in different layers of deep models over three event cases. (a) Goma-VE-2021. (b) Beirut-EP-2023. (c) Libya-FL-2023. (d) Hawii-WF-2023. (e) Turkey-EQ-2023. (f) Acapulco-HC-2023. In the visualization, closer to red indicates larger response values, and closer to dark blue the opposite. See Appendix \ref{app:feat_visualization} for implementation details.}

  \label{fig:cam_different_events}
\end{figure*}

\par To better understand the models' behavior beyond performance metrics, we explore the internal attention patterns of the trained ChangeOS using class activation maps (CAMs) \citep{Selvaraju2017Grad}. Figure \ref{fig:cam_different_events} presents the CAM responses of ChangeOS across six representative disaster events: volcano, explosion, flood, wildfire, earthquake, and hurricane. We observe that the attention distribution varies across disaster types. Taking Goma-VE-2021 as an example, for the “Destroyed” category, the encoder exhibits strong activations in nearly all the built-up regions in the optical images, accurately localizing individual buildings. This suggests that the model has effectively learned to extract detailed structural cues from pre-event optical imagery. In contrast,  for the SAR images, the encoder shows intense activation over the lava-covered regions on the left. This indicates that the model has identified the lava-covered regions as a key signal for destruction, likely due to the significant backscatter changes caused by lava flow. In the “Damaged” category, the activations are more subtle. Attention is primarily focused near the boundary of the lava flow, where partial or ambiguous structural changes occur. In the decoder, the destroyed buildings in the lava-affected area are strongly activated, which aligns well with the reference labels. Conversely, for the “Damaged” class, only a few regions are activated. This suggests that inferring partial damage from SAR imagery in volcanic disaster scenarios remains a significant challenge, as subtle structural degradation is often not clearly reflected in SAR backscatter or texture. 

\begin{figure*}[!t]
    \centering
    \includegraphics[width=5.5in]{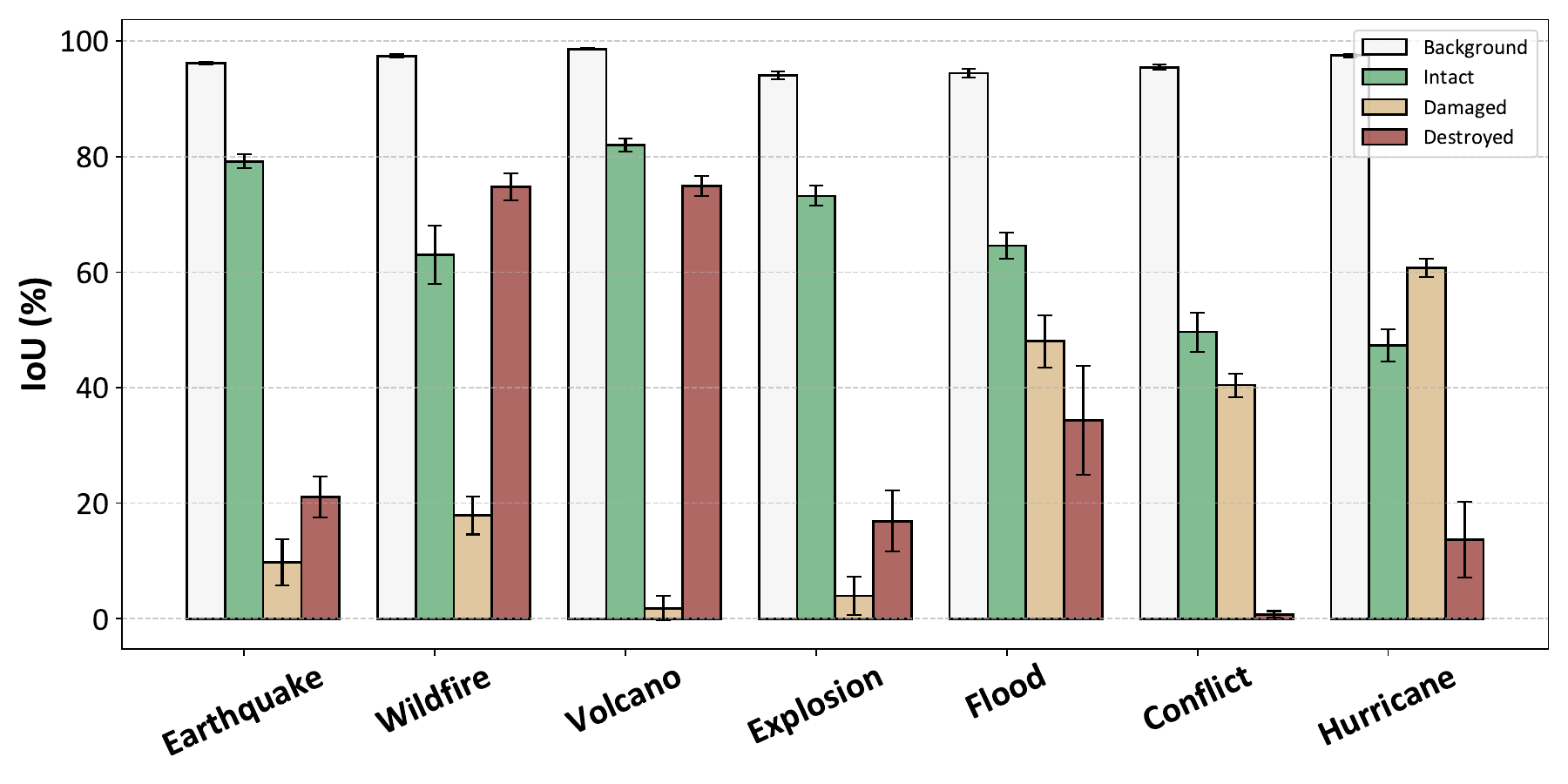}
   \caption{IoU distribution of deep models over seven disaster types. Each bar represents the average IoU of seven DL models for that specific category under each disaster type. The error bars indicate the standard deviation of IoU scores across the seven models.}
  \label{fig:iou_distribution_over_diasster}
\end{figure*}

\begin{figure*}[!t]
    \centering
    \includegraphics[width=6.5in]{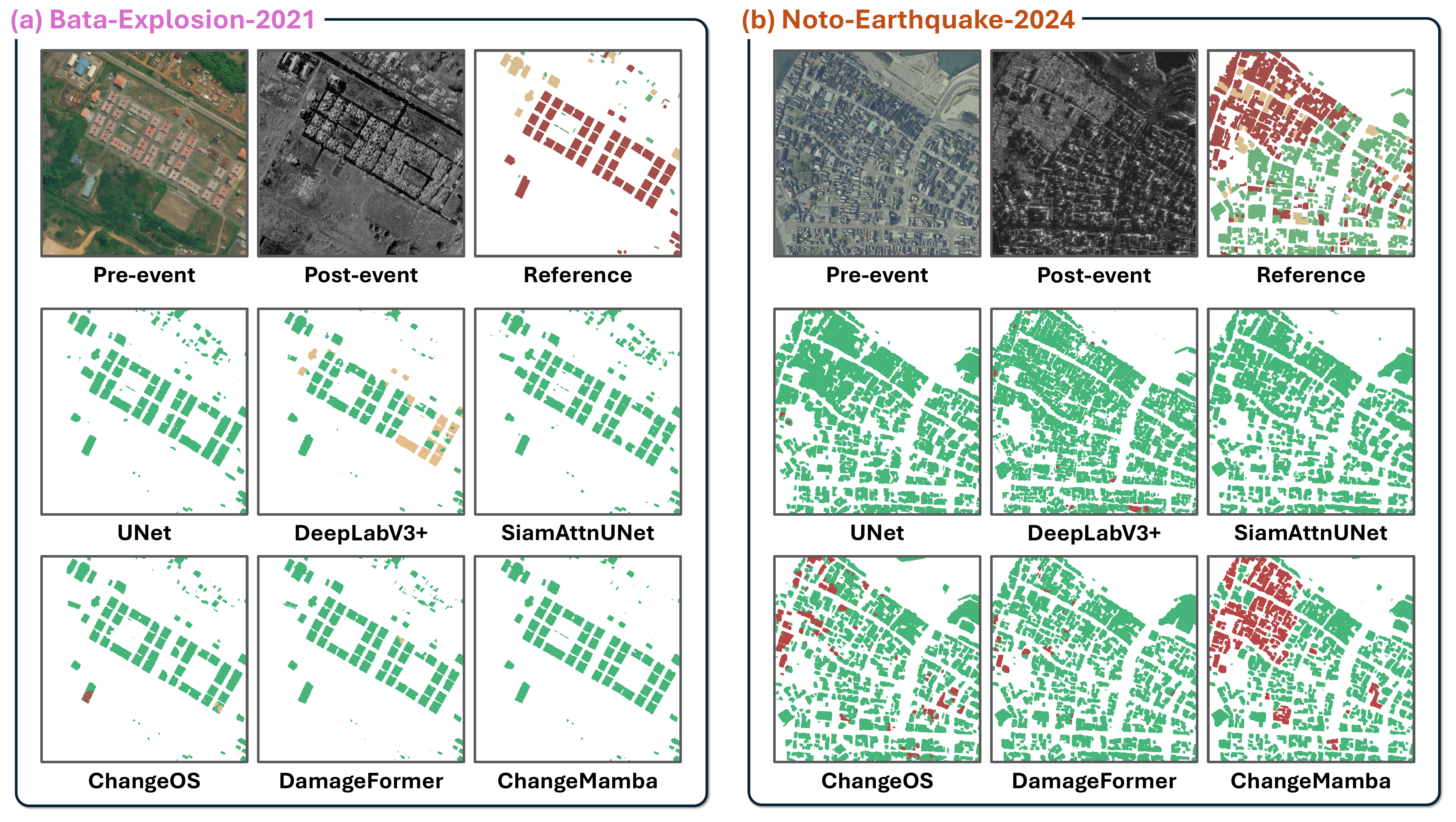}
   \caption{Typical failure cases of different models on Bata-Explosion-2021 and Noto-Earthquake-2024 in \textsc{Bright}, where optical images are from © Maxar and © GSI Japan and SAR images are from © Capella Space and © Umbra. }
  \label{fig:failure_cases}
\end{figure*}

\par Beyond understanding what the models have already learned, a more critical question is: What can they learn? More specifically, to what extent can optical–SAR modality improve the accuracy of building damage assessment across diverse disaster scenarios? To investigate this, we aggregate the IoUs of the seven models and calculate their average IoU across the seven major disaster types, as shown in Figure \ref{fig:iou_distribution_over_diasster}. The wildfire and volcano events exhibit the highest IoU scores for the “Destroyed” category, both exceeding 70$\%$. This indicates that the model can effectively leverage SAR-based backscatter anomalies, such as lava flows or widespread debris fields, to detect fully destroyed structures. These results demonstrate a strong potential of optical–SAR fusion in such high-impact scenarios. For the “Damaged” category, the performance drops significantly. The average IoU for damaged buildings in wildfire events remains below 20$\%$, while in volcano events, it falls to around 5$\%$. This suggests that single-polarization SAR imagery lacks the fine-grained information needed to reliably distinguish partially damaged buildings, where structural integrity may still be partially preserved and backscatter signals remain ambiguous. In the earthquake events, both damaged and destroyed categories yield relatively low IoUs. This is likely due to the complex and heterogeneous patterns of structural collapse typical of seismic events, where damage is often subtle, partial, and highly variable. These conditions pose significant challenges for SAR-based assessment. Interestingly, the model achieves relatively high IoU scores in the flood and hurricane events for the “Damaged” category with approximately 50$\%$ and 60$\%$, respectively. This indicates that SAR effectively captures contextual environmental changes, such as water inundation or terrain disruption, which indirectly aid in assessing building damage. In the case of the conflict event, the model's performance on the “Destroyed” class is surprisingly low. This might be attributed to the limited number of destroyed samples in the dataset for this category, which leads to insufficient learning and poor generalization.

\par These quantitative limitations are vividly illustrated by the typical failure cases shown in Figure \ref{fig:failure_cases}. In the Bata-Explosion-2021 event, models misclassify severely destroyed buildings as intact, reflecting the difficulty of interpreting heterogeneous debris patterns. Similarly, in the Noto-Earthquake-2024 event, large-scale collapses are largely missed, highlighting the challenge of diverse and subtle seismic damage. These examples visually confirm that the significant heterogeneity in damage patterns makes it challenging for models to learn a consistent and generalizable representation of damage.

\par In summary, these findings confirm both the promise and limitations of optical-SAR modality for all-weather, global-scale disaster response. Although this combination performs well in events characterized by large-scale surface disruption (e.g., wildfires, volcanoes), it struggles with subtle or localized damage patterns. Incorporating richer data sources, such as fully polarimetric SAR and LiDAR data, can further enhance the accuracy and reliability of future all-weather building damage assessments.

\begin{table*}[!t]
  \renewcommand{\arraystretch}{1.25}
  \caption{Performance comparison of UNet and DeepLabV3+ using only post-event SAR input and pre-event optical plus post-event SAR inputs for damage classification task. Here, accurate building masks are provided as the post-processing step to all models to isolate the effect of building localization task on the damage classification task.}
  \label{tbl:role_of_optical_data}
  \centering
  \begin{tabular}{l  | l | c c c c c c}
    \hline	
  \multirow{2}{*}{Method} & \multirow{2}{*}{Modality} & \multirow{2}{*}{$F_{1}^{clf}$ ($\%$)}   & \multirow{2}{*}{Final mIoU ($\%$)}  & \multicolumn{4}{c}{IoU per class ($\%$)}  \\
  \cline{5-8} 
     &  & & & Background  &  Intact &  Damaged & Destroyed  \\
  \hline \hline
    \multirow{2}{*}{UNet} & Post-event SAR &  68.71  & 69.84 & 100.0 & 88.19 &  35.83 & 55.35 \\
      & Pre-event optical + post-event SAR & 73.59  &  72.41 & 100.0 & 89.38 &  44.83 & 55.42 \\
     \hline
    \multirow{2}{*}{DeepLabV3+} & Post-event SAR & 72.12  &  72.19 & 100.0 & 89.59 & 39.63 & 59.54 \\
    & Pre-event optical + post-event SAR & 73.90  & 73.93 & 100.0 & 90.32 &  40.45 & 64.94 \\
    \hline
\end{tabular}
\end{table*}

\begin{table*}[!t]
  \renewcommand{\arraystretch}{1.25}
  \caption{Performance comparison of different post-event modalities on a subset of \textsc{Bright}. Results are reported for UNet, DeepLabV3+, and DamageFormer on five disaster events where high-quality post-event optical imagery is available: Bata-Explosion-2020, Beirut-Explosion-2021, Hawaii-Wildfire-2023, Libya-Flood-2023, and Noto-Earthquake-2024.}
  \label{tbl:different_mode_comparison}
  \centering
  \begin{tabular}{l  | l | c c c c c c c}
    \hline	
  \multirow{2}{*}{Method} & \multirow{2}{*}{Post-event modality} & \multirow{2}{*}{$F_{1}^{loc}$ ($\%$)} & \multirow{2}{*}{$F_{1}^{clf}$ ($\%$)}   & \multirow{2}{*}{Final mIoU ($\%$)}  & \multicolumn{4}{c}{IoU per class ($\%$)}  \\
  \cline{6-9} 
     &  & & & & Background  &  Intact &  Damaged & Destroyed  \\
  \hline \hline
    \multirow{3}{*}{UNet} & SAR & 85.05 & 71.43  & 62.94 & 94.39 & 65.60 &  42.34 & 49.43 \\
    & Optical & 86.46 & 75.64  &  65.96 & 94.56 & 68.62 &  45.27 & 55.36 \\
      & Optical+SAR & 86.70 & 74.46  &  66.29 & 94.72 & 69.16 &  41.73 & 59.57 \\
    \hline
    \multirow{3}{*}{DeepLabV3+} & SAR &  83.55 & 67.52  & 60.57 & 93.86 & 65.62 & 35.64 & 47.16  \\
    & Optical & 85.79 & 74.39  & 64.87 &94.33 & 67.90 & 44.31 & 52.94  \\
    & Optical+SAR & 85.90 & 74.87  & 65.84 &  94.48 & 69.60 &  44.68 & 54.60 \\
     \hline
    \multirow{3}{*}{DamageFormer} & SAR &  88.41 & 73.43 & 65.56 & 95.30 &  70.62 & 41.31 & 55.00 \\
    & Optical  & 88.32 & 78.04 & 69.76 &  95.37 &  72.72 & 47.26 & 63.68 \\
    & Optical+SAR  & 88.86 & 79.27 & 70.79 &  95.56 &  73.64 & 48.44 &  65.51 \\
    \hline
\end{tabular}
\end{table*}
\subsection{The role of optical pre-event data in multimodal building damage assessment}\label{sec:role_of_optical}

\par In the last section, CAM visualizations revealed that DL models also exhibit responses to disaster-specific patterns in pre-event optical imagery. This observation suggests that optical data may play a more complex role in multimodal building damage mapping than simply supporting building localization. In other words, in a multimodal bi-temporal setup, does pre-event optical imagery act solely as a localization aid, or does it provide additional semantic cues that networks can exploit for more accurate damage classification?

\par To explore this, we conducted controlled experiments using UNet and DeepLabV3+. Both networks were trained under two configurations: (i) using post-event SAR imagery only, and (ii) using multimodal pre- and post-event inputs (optical-SAR). To isolate the contribution of pre-event optical data beyond building localization, we provided perfect building masks for postprocessing in both settings. This design ensures that any observed differences in performance are attributable to the additional information from pre-event optical imagery, rather than differences in network architecture or localization accuracy.

\par The results, summarized in Table~\ref{tbl:role_of_optical_data}, show that incorporating pre-event optical imagery leads to notable improvements in distinguishing building damage levels. For UNet, the IoU for the ``Damaged'' class increased from 35.83\% (SAR only) to 44.83\% (Optical-SAR), and for the ``Destroyed'' class from 55.35\% to 55.42\%. DeepLabV3+ exhibited significant gains also, with IoU improvements from 39.63\% to 40.45\% for ``Damaged'' category, and from 59.54\% to 64.94\% for ``Destroyed'' category. These results suggest that pre-event optical imagery contributes beyond mere building localization, enriching the feature space for more effective semantic comparison for different building damage levels across modalities.

\subsection{Impact of post-event modality on building damage assessment performance}

\par Although the primary design of \textsc{Bright} is to facilitate all-weather disaster response through the use of pre-event optical and post-event SAR imagery, it is also important to understand how these modalities compare when high-quality post-event optical imagery is available. To this end, we conducted supplementary experiments on a subset of events, including Bata-Explosion-2020, Beirut-Explosion-2021, Hawaii-Wildfire-2023, Libya-Flood-2023, and Noto-Earthquake-2024, for which pre-processed post-event optical data were accessible. We evaluated three experimental setups: (i) optical-only (pre-event optical + post-event optical), (ii) SAR-only (pre-event optical + post-event SAR, i.e., the standard \textsc{Bright} setting), and (iii) optical+SAR fusion (pre-event optical + post-event optical + post-event SAR). 

\par Table~\ref{tbl:different_mode_comparison} presents the experimental results. As expected, when ideal post-event optical imagery is available, the optical-only setup achieves higher performance than the SAR-only setup. For example, with DamageFormer, the optical-only configuration reaches a final mIoU of 69.76\%, compared to 65.56\% for SAR-only.  Importantly, the performance gap between optical and SAR is not substantial, demonstrating that SAR alone provides a strong alternative in the absence of usable optical imagery. Moreover, the fusion of optical and SAR consistently yields the best results across all tested models. For instance, DamageFormer’s mIoU further increases to 70.79\% with Optical+SAR fusion, indicating that SAR contributes complementary information that strengthens performance even under optimal optical conditions. 

\par These findings underscore two important insights. First, multimodal fusion is beneficial even when high-quality optical data are available, as SAR provides unique structural information that enriches the optical signal. Second, the performance of the SAR-only approach, being reasonably close to the optical-only results, highlights the practical value of SAR in real-world disaster scenarios where post-event optical imagery is often unavailable. \textsc{Bright} is therefore designed to advance the development of models for these realistic, often non-ideal, but operationally critical all-weather disaster response settings.

\subsection{Effect of post-processing method}\label{sec:pps}
\newcommand{\upaccuracy}[2]{
#1 \fontsize{7pt}{1em}\selectfont\color{orange}{$\!\uparrow\!$ \textbf{#2}}
}
\newcommand{\downaccuracy}[2]{
#1 \fontsize{7pt}{1em}\selectfont\color{cyan}{$\!\downarrow\!$ \textbf{#2}}
}
\begin{table}[!t]
  \renewcommand{\arraystretch}{1.25}
  \caption{Further contributions to mIoU from post-processing algorithms. ChangeMamba \citep{Chen2024ChangeMamba} is used here as the baseline. Details on these algorithms are provided in Appendix \ref{app:pps}.}
  \label{tbl:pps_methods}
  \centering
  \begin{tabular}{l | l l }
    \hline	
  Method &  mIoU (set) &  mIoU (event)  \\
    \hline
    \hline
    Baseline & 67.63  & 51.39\\
     \hline
    Test-time augmentation  & \upaccuracy{68.50}{0.87}  & \upaccuracy{51.95}{0.56}\\
    Object-based majority voting & \downaccuracy{67.22}{0.41}  & \upaccuracy{52.08}{0.69}  \\
    Ensembling multiple models  & \upaccuracy{68.45}{0.82}  &  \upaccuracy{52.14}{0.75} \\
    \hline
    All &  \upaccuracy{68.86}{1.23} & \upaccuracy{52.31}{0.92} \\
   
    \hline
\end{tabular}
\end{table}
\par Post-processing techniques help refine raw predictions from DL models, to reduce noise, improve consistency, and ensure spatial coherence in damage maps \citep{ZHENG2021Building}. Here, we explore the impact of post-processing algorithms. Table \ref{tbl:pps_methods} presents the effect of three post-processing techniques applied to ChangeMamba. The post-processing methods evaluated include test-time augmentation, object-based majority voting, and model ensembling. The details of the methods are provided in Appendix \ref{app:pps}. As shown in Table \ref{tbl:pps_methods}, the test-time augmentation improves the mIoU by 0.87$\%$ at the set level and 0.56$\%$ at the event level, demonstrating its effectiveness in enhancing model robustness across diverse disaster scenarios. Object-based majority voting, which aggregates predictions at the building-object level to enforce spatial consistency, slightly reduces set-level mIoU (-0.41$\%$) but improves event-level mIoU (+0.69$\%$). Ensembling multiple models leads to a 0.82$\%$ increase in mIoU at the set level and a 0.75$\%$ increase at the event level, reinforcing its effectiveness in improving model performance across different disaster events. Applying all post-processing techniques together yielded the highest performance improvement, with a 1.23$\%$ increase in set-level mIoU and a 0.92$\%$ increase in event-level mIoU. These results confirm that combining different post-processing methods can significantly enhance the reliability of AI-based damage assessments, ensuring better generalization across disaster types and locations. In summary, post-processing techniques are crucial in improving the accuracy of building damage maps. Future work can further explore adaptive post-processing strategies tailored to specific disaster types to enhance prediction reliability in multimodal EO data contexts.

\begin{table*}[!t]
  \scriptsize \renewcommand{\arraystretch}{1.65}
  \caption{The mIoU on different events for different DL models in zero-shot and one-shot cross-event transfer setups. The highest values are highlighted in \textbf{\textcolor{purple}{purple}}, and the second-highest results are highlighted in \textbf{\textcolor{teal}{teal}}.}
  \label{tbl:bda_BRIGHT_event_zero_one_shot}
  \centering
  \begin{tabular}{l | c c c | c c c c}
    \hline	
   \multirow{2}{*}{Events} & UNet  & DeepLabV3+  & SiamAttnUNet  & SiamCRNN   & ChangeOS   & DamageFormer  & ChangeMamba  \\
   \cline{2-8}
   & zero- / one-shot & zero- / one-shot & zero- / one-shot & zero- / one-shot & zero- / one-shot & zero- / one-shot & zero- / one-shot \\
    \hline
    \hline
     Beirut-EP-2020  & 31.62 / 34.89 &  34.72 / 33.58	 & 35.33 / 29.17 &  \textbf{\textcolor{teal}{39.43}} / \textbf{\textcolor{purple}{40.94}} & 38.48 / 38.21  & 38.71 / 39.93 & \textbf{\textcolor{purple}{40.30}} / \textbf{\textcolor{teal}{40.92}} \\ 
     
     Bata-EP-2021   & 33.48 / 37.48 & 33.55 / 30.23 & 33.78 / 35.24 & 37.46 / \textbf{\textcolor{teal}{40.06}} &	38.00 / 38.18  & \textbf{\textcolor{teal}{38.91}} / 39.42  & \textbf{\textcolor{purple}{39.17}} / \textbf{\textcolor{purple}{40.31}} \\ 
     
     Goma-VE-2021   & 55.97 / 57.14 &  55.01 / 58.13  & 57.52 / 55.75 & \textbf{\textcolor{teal}{58.91}} / 59.54 & 58.66	/ 57.85  &	55.86 /  \textbf{\textcolor{teal}{59.89}} & \textbf{\textcolor{purple}{60.39}} / \textbf{\textcolor{purple}{61.91}} \\ 
     Les Cayes-EQ-2021 & 26.21 / 38.72 & 35.50 / 37.90 & 37.22 / 38.76 & 39.45 / 40.28  &	39.36 / 40.26 &	\textbf{\textcolor{teal}{40.41}} / \textbf{\textcolor{teal}{41.07}}  & \textbf{\textcolor{purple}{41.35}} / \textbf{\textcolor{purple}{42.06}} \\ 
     La Palma-VE-2021 & 32.74 / 35.29 & 30.86 / \textbf{\textcolor{teal}{36.25}} & \textbf{\textcolor{teal}{34.00}} / 34.99 & 32.90 / 33.32 &	29.96 / 32.60  & 32.28 / \textbf{\textcolor{purple}{37.62}}  & \textbf{\textcolor{purple}{34.18}} / 34.93 \\ 
     Marshall-WF-2021  &  34.97 / 37.84 & 35.05 / 37.12 & 35.23 / 39.54 & 39.82 / \textbf{\textcolor{teal}{46.83}} &	36.31 / 43.72  & \textbf{\textcolor{purple}{40.93}} / \textbf{\textcolor{purple}{51.15}} & \textbf{\textcolor{teal}{40.61}} / 45.96 \\ 
     Ukraine-AC-2022 & 32.36 / 37.50  & 32.23 / 35.98 & 33.84 / 35.76 & 34.77 / \textbf{\textcolor{teal}{38.14}} &	\textbf{\textcolor{purple}{36.30}} / \textbf{\textcolor{purple}{39.28}}  & 33.84 / 36.51 & \textbf{\textcolor{teal}{35.78}} / 36.99 \\ 
     Turkey-EQ-2023  & 37.82 / 35.83 &  37.28 / \textbf{\textcolor{teal}{40.49}} & 36.09 / 37.49 & 38.32 / 40.11 &	\textbf{\textcolor{teal}{38.45}} / 39.51  & 37.20 / 40.27  & \textbf{\textcolor{purple}{41.28}} / \textbf{\textcolor{purple}{42.80}} \\  
     Kyaukpyu-CC-2023  & 31.46 / 31.71 & 34.01 / 34.75 & 33.77 / 35.93 & \textbf{\textcolor{teal}{34.19}} / \textbf{\textcolor{purple}{37.71}} & \textbf{\textcolor{purple}{36.66}} / \textbf{\textcolor{teal}{36.25}}  & 32.99 / 33.54  &  33.85 / 36.08 \\ 
     Hawaii-WF-2023  &  31.73 / 39.80 & 31.32 / 45.44 & \textbf{\textcolor{teal}{31.79}} / 45.12 & 30.94 / 42.65 &  31.15 / 41.96	&	 31.60  / \textbf{\textcolor{teal}{50.64}} & \textbf{\textcolor{purple}{36.14}} / \textbf{\textcolor{purple}{50.89}} \\ 
     Morocco-EQ-2023  & 38.24 / 39.81 & 40.48 / 41.23 & 40.78 / 40.69 & 40.89 / 42.89 &	41.40 / 41.89  & \textbf{\textcolor{teal}{41.89}} / \textbf{\textcolor{teal}{43.42}}  &  \textbf{\textcolor{purple}{42.86}} / \textbf{\textcolor{purple}{43.51}} \\ 
     Derna-FL-2023  & 33.25 / 38.44 & 35.07 / 40.02  & 33.76 / 42.15 & 35.88 / \textbf{\textcolor{teal}{44.77}} &	35.80 / 43.85  & \textbf{\textcolor{teal}{36.27}} / 44.69  & \textbf{\textcolor{purple}{37.02}} / \textbf{\textcolor{purple}{45.47}} \\ 
     Acapulco-HC-2023 & 26.99 / 38.76 & 27.78 / 33.24 & \textbf{\textcolor{purple}{30.93}} / \textbf{\textcolor{teal}{38.94}} & 26.70 / 34.53 &	27.06 / 34.50  & 27.48 / 32.87 & \textbf{\textcolor{teal}{28.45}} / \textbf{\textcolor{purple}{38.96}} \\ 
     Noto-EQ-2024 & 31.02 / 33.83 & 35.57 / 39.17 & 35.87 / 36.22 & 36.04 / 38.52&	\textbf{\textcolor{teal}{39.14}} / 39.15  & 36.37 / \textbf{\textcolor{teal}{43.93}} &  \textbf{\textcolor{purple}{39.90}} / \textbf{\textcolor{purple}{44.33}} \\ 
    \hline
    Average & 34.13 / 38.36 & 35.60 / 38.82 & 36.42 / 38.98 & 37.55 / 41.45 &	 \textbf{\textcolor{teal}{37.62}} / 40.52 &	37.48 / \textbf{\textcolor{teal}{42.49}} & \textbf{\textcolor{purple}{39.38}} / \textbf{\textcolor{purple}{43.23}} \\ 
    \hline
\end{tabular}
\end{table*}

\subsection{Evaluation on cross-event transfer setup}

\subsubsection{Baseline methods}
Cross-event transfer, especially under zero-shot settings, poses a significant challenge for building damage assessment. As shown in Table \ref{tbl:bda_BRIGHT_event_zero_one_shot}, the average mIoU of all baseline models in the zero-shot setting is below 40$\%$, a noticeable drop compared to their performance under the fully supervised standard ML data split in Table~\ref{tbl:bda_BRIGHT_event}, where the models typically achieve 48$\%$ to 52$\%$ mIoU. This performance gap underscores the difficulty of generalizing to unseen disaster events without access to any target domain supervision due to substantial domain shifts in imaging conditions, damage patterns, urban morphology, and sensor response. Despite this, all models exhibit clear performance gains in the one-shot setting, where a small number of labeled samples from the target event are available. This suggests that even minimal supervision can significantly aid adaptation to new disaster contexts. 

\par Among the evaluated models, ChangeMamba consistently achieves the highest overall performance, with an average mIoU of 39.38$\%$ in zero-shot and 43.23$\%$ in one-shot settings, followed by DamageFormer. This highlights the strength of recent Transformer- and Mamba-based architectures in transferring learned knowledge under complex multimodal and disaster scenarios. Models using a decoupled architecture generally outperform direct segmentation models, which confirms that separating building localization and damage classification improves generalization in cross-event transfer tasks. In contrast, the Acapulco-HC-2023 and Ukraine-AC-2022 events show weaker results across all models, reflecting the difficulty in transferring to domains with limited or inconsistent destruction patterns.

\subsubsection{Unsupervised domain adaptation and semi-supervised learning methods}

\begin{table}[!t]
  \renewcommand{\arraystretch}{1.25}
  \caption{Results of unsupervised domain adaptation methods adopted for zero-shot cross-event transfer setup. DeepLabV3+ \citep{Chen2018Encoder} is used as the baseline here. Detailed mIoU on each event is listed in Table \ref{tbl:detailed_mIoU_UDA}.}
  \label{tbl:UDA_methods}
  \centering
  \begin{tabular}{l | l }
    \hline	
  Method &  Avg. mIoU  \\
    \hline
    \hline
    Source-Only & 35.60 \\
     \hline
    AdaptSeg \citep{Tsai2018Learning} & \upaccuracy{36.05}{0.45}  \\
    AdvEnt \citep{Vu2019ADVENT} &  \downaccuracy{35.43}{0.17}  \\
    CLAN \citep{Luo2019Taking}  & \upaccuracy{36.01}{0.41}  \\
    PyCDA \citep{Lian2019Constructing}  & \downaccuracy{33.19}{2.41} \\
    FDA \citep{Yang2020FDA}  &   \upaccuracy{36.29}{0.69} \\
    \hline
\end{tabular}
\end{table}

\begin{table}[!t]
  \renewcommand{\arraystretch}{1.25}
  \caption{Results of semi-supervised learning methods adopted for one-shot cross-event transfer setup. DeepLabV3+ \citep{Chen2018Encoder} is used as the baseline here. Detailed mIoU on each event is listed in Table \ref{tbl:detailed_mIoU_SSL}. }
  \label{tbl:SSL_methods}
  \centering
  \begin{tabular}{l   | l}
    \hline	
  Method &  Avg. mIoU \\
    \hline
    \hline
    Baseline &  38.82 \\
     \hline
     MT \citep{Guyon2017Mean} &  \upaccuracy{40.00}{1.18}  \\
     CCT \citep{Ouali2020Semi} &  \upaccuracy{39.49}{0.67} \\
     GCT \citep{Ke2020Guided} & \upaccuracy{39.34}{0.52}  \\
     CPS \citep{Chen2021Semi} & \downaccuracy{37.06}{1.76} \\
    \hline
\end{tabular}
\end{table}

\par In the context of cross-event transfer, the UDA and SSL methods naturally emerge as promising strategies to bridge the domain gap between the source and the target disaster events. In the zero-shot setting, UDA methods aim to improve model generalization by aligning the source and target domains without requiring any target labels. In the one-shot setting, SSL methods leverage a small number of labeled samples with abundant unlabeled data from the target event, making them especially appealing for real-world disaster scenarios where rapid and comprehensive annotation is infeasible. It remains an open question whether these UDA and SSL methods, originally developed for natural image domains, can effectively handle the challenges of multimodal EO data in complex disaster scenarios. We evaluated several representative UDA and SSL methods using DeepLabV3+ as the baseline model to examine their capabilities in complex disaster scenarios. The results for UDA and SSL are reported in Tables \ref{tbl:UDA_methods} and \ref{tbl:SSL_methods}, respectively.

\par Table \ref{tbl:UDA_methods} presents the performance of five classical UDA methods. Compared to the source-only baseline (35.60$\%$ mIoU), most methods achieve modest improvements, with FDA (+0.69$\%$) and AdaptSeg (+0.45$\%$) showing the most consistent gains. CLAN also performs slightly better (+0.41$\%$), suggesting that category-level alignment contributes positively even under a large domain shift. In contrast, AdvEnt shows negligible change, and PyCDA significantly underperforms, dropping 2.41$\%$ mIoU below the baseline. This performance degradation indicates that approaches relying on pseudo-label refinement or curriculum learning may struggle in multimodal imagery under disaster scenarios, where spatial layout and damage semantics vary drastically across events. Overall, while UDA methods show some promise, their improvements are relatively minor and not robust across all events, underscoring the difficulty of domain alignment in EO-based damage mapping.

\begin{figure*}[!t]
   \centering
    \includegraphics[width=6.9in]{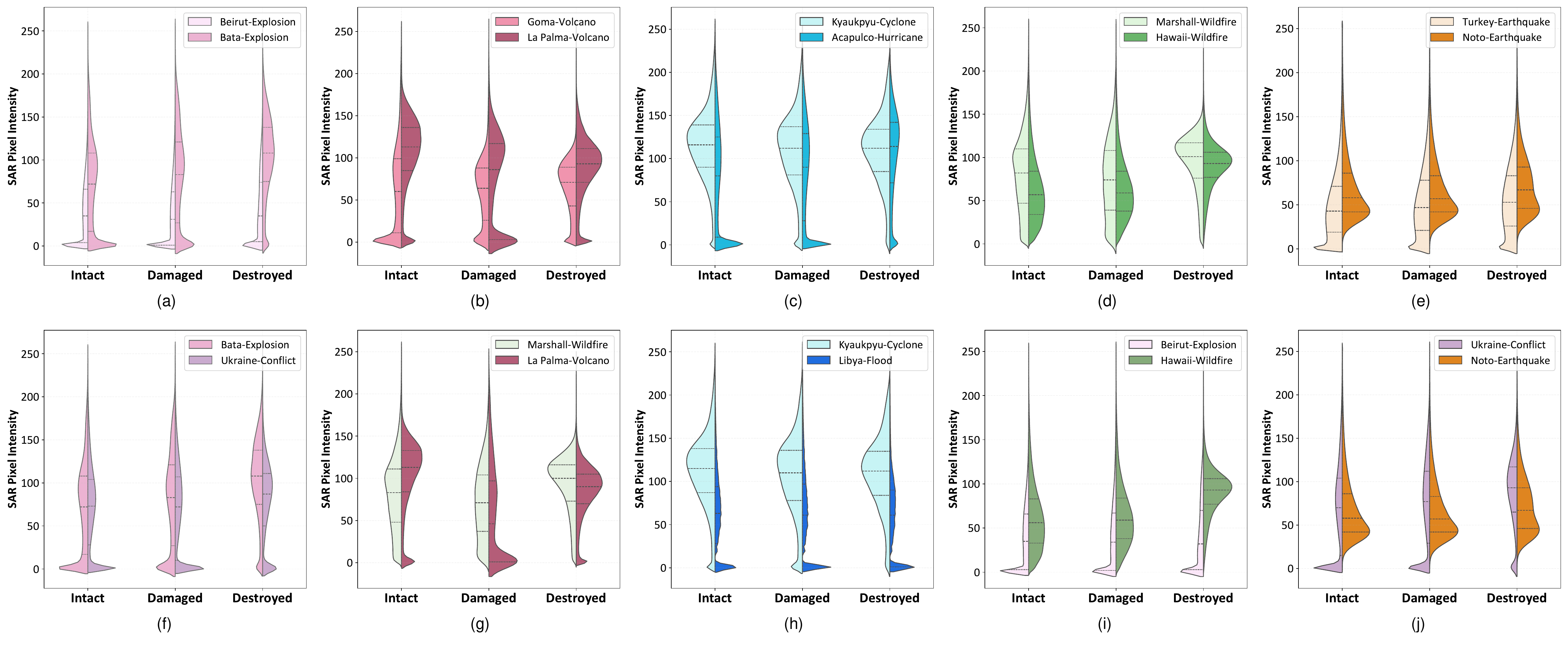}
   \caption{The comparison of pixel distribution of different categories in SAR images for different events. The first row compares different events with the same disaster type. (a) Explosion. (b) Volcano eruption. (c) Wildfire. (d) Cyclone/Hurricane (e) Earthquake. The second row compares different disaster types. (f) Explosion \emph{vs} Conflict. (g) Wildfire \emph{vs} Volcano Eruption. (h) Cyclone \emph{vs} Flood. (i) Explosion \emph{vs} Wildfire. (j) Conflict \emph{vs} Earthquake.}

  \label{fig:pixel_distribution_same_event}
\end{figure*}

\par In the one-shot setting, we evaluated several popular SSL methods, as shown in Table \ref{tbl:SSL_methods}. With only minimal supervision in the target event, all methods except CPS improved over the one-shot baseline (38.82$\%$). The best-performing method is Mean Teacher (MT), which yields a gain of +1.18$\%$, followed by CCT (+0.67$\%$) and GCT (+0.52$\%$). These results show that simple consistency-based teacher-student frameworks are particularly effective in leveraging unlabeled data under limited supervision, likely due to their robustness to noisy or class-imbalanced targets. 

\subsubsection{Why is cross-event transfer challenging?}

\par To better understand why cross-event generalization is difficult, we explore two fundamental factors rooted in the nature of real-world disaster response.

\begin{enumerate}
    \item \textbf{Inconsistent damage signatures across events}. Figure \ref{fig:pixel_distribution_same_event} presents violin plots of SAR backscatter values for intact, damaged, and destroyed buildings across multiple event pairs. These plots reveal two key observations. First, even within the same disaster category (the first row), the pixel intensity distributions for damaged and destroyed buildings differ significantly between events. This indicates that SAR-based damage signatures are inconsistent across locations, possibly due to differences in urban layout, building materials, or sensor incidence angles. Secondly, the distribution shift becomes more pronounced across different types of disasters (the second row). For example, the signatures of destroyed buildings in wildfires notably differ from those in volcano eruptions, which also differ from floods or hurricanes. These variations reflect fundamental differences in damage mechanisms: buildings burned in wildfires, submerged in floods, or collapsed in earthquakes leave very different patterns in SAR backscatter. Such distributional discrepancies make it extremely difficult for models to generalize from one event to another. A model trained on one disaster might learn class boundaries (e.g., between “Damaged” and “Destroyed”) that do not transfer well to another disaster, especially when the visual and physical properties of damage are fundamentally different.
\begin{figure}[!t]
        \centering
        \includegraphics[width=3.3in]{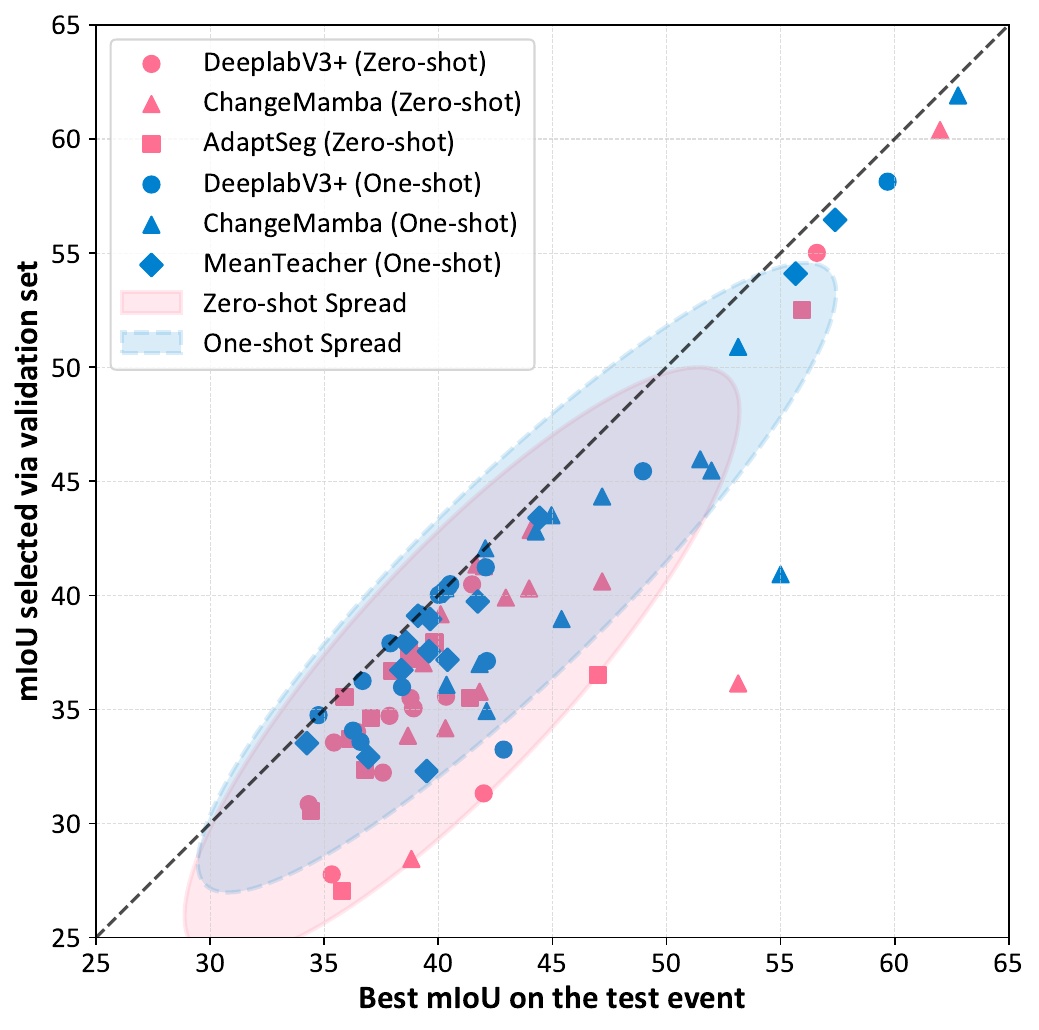}
       \caption{Comparison of models' best performance (mIoU) on test events versus the best checkpoints selected on validation sets under the cross-event transfer setting. Each point represents the performance on a single test event under cross-event transfer. The farther a point lies from the diagonal line, the larger the gap between the model's selected performance and its true upper bound.}
      \label{fig:model_selection_vs_oracle_accuracy}
\end{figure}

    \item \textbf{Lack of target domain supervision for model selection}. Another critical challenge in cross-event setup is the absence of labeled target domain samples for model selection. In typical domain adaptation benchmarks in computer vision, a validation set of target domain is available to tune hyperparameters or select the best checkpoint. However, no such validation data can be assumed in disaster response scenarios. Figure \ref{fig:model_selection_vs_oracle_accuracy} illustrates the resulting issue: the gap between the actual best mIoU on the test event and the mIoU of the checkpoint selected using a source domain validation set for several representative models under both zero-shot and one-shot settings. In the zero-shot setting, this performance gap is substantial across most models. This indicates that relying on source domain validation leads to suboptimal model selection due to poor reflection of the target distribution. The one-shot setting helps reduce this gap by enabling limited target-aware selection. However, the issue is not entirely resolved. Even with a few labeled samples, model instability and domain shift still make selection challenging.
\end{enumerate}

\par Together, these two findings highlight that cross-event transfer is hindered not only by semantic and statistical shifts across disaster types but also by operational constraints that prevent ideal training and tuning. Future studies should consider both aspects, \emph{i.e.}, developing models that are robust to distributional variance and designing selection strategies that do not depend on target supervision, such as self-validation, early stopping heuristics, or domain-agnostic performance proxies \citep{yang2023can}.

\subsection{Evaluation of unsupervised multimodal change detection methods}

\par Unsupervised multimodal change detection (UMCD) plays a crucial role in post-disaster assessment by enabling rapid analysis of affected areas without requiring labeled data or prior model training \citep{Lv2022Land}. Current studies are mainly conducted on toy datasets with limited geographic diversity and scene complexity. These datasets are for general land cover changes. Whether these methods work in real-world disaster occurrence/building damage scenarios is still unknown. \textsc{Bright} offers a new opportunity to evaluate UMCD methods to provide insights into their robustness and scalability in disaster response scenarios. Six representative UMCD methods \citep{Sun2021a, chen2022unsupervised, Chen2023Fourier, Han2024Unsupervised, Sun2024Image, Liu2025AEKAN} are evaluated in this work. The experimental setup is described in Appendix \ref{app:UMCD}.

\begin{table*}[!t]
  \renewcommand{\arraystretch}{1.25}
  \caption{Results of representative unsupervised multimodal change detection methods. KC is the acronym of kappa coefficient. The highest values are highlighted in \textbf{\textcolor{purple}{purple}}, and the second-highest results are highlighted in \textbf{\textcolor{teal}{teal}}. The accuracies on the UMCD benchmark dataset are the accuracies on the four datasets presented in Figure \ref{fig:UMCD_benchmark_dataset}, obtained from their literature. Details of methods and benchmark datasets are presented in Appendix \ref{app:UMCD}. The random guessing baseline is included to indicate the performance floor under the UMCD setup. The ``-'' symbol indicates that the corresponding method did not report results on that dataset in their original publications.}
  \label{tbl:UMCD_evlaution}
  \centering
  \begin{tabular}{l | c c c | c c c c }
    \hline	
  \multirow{2}{*}{Method} & \multicolumn{3}{c}{UMCD benchmark datasets} &  \multicolumn{4}{c}{\textsc{Bright}}   \\
  \cline{2-8}
  &  OA & F1 & KC  &  OA & F1 & IoU & KC \\
    \hline
    \hline
    \rowcolor{gray!13} \emph{Random} &  50.0 & 8.4 / 6.0 / 11.0 / 11.4 & 0.0 & 50.00 & 7.83 & 4.08 & 0.00 \\
    IRG-McS \citep{Sun2021a} & 98.3 / - / 97.1 / 97.2 & 80.4 / - / 75.4 / 73.7 &   79.4 / - / 73.9 / 75.1 &  \textcolor{purple}{\textbf{90.03}} & 12.65 & 6.75 & 7.65 \\
    SR-GCAE \citep{chen2022unsupervised} & 98.6 / 98.5 / - / - & 82.9 / 77.6 / -  / - & 82.1 / 76.9 / - / -  & 77.83 &  14.35 & 7.73 &  5.64  \\
    FD-MCD \citep{Chen2023Fourier} & 98.2 / 97.8 / - / 96.7 &  81.4 / 72.2 / - / 73.2 & 82.3 / 71.1 / - / 71.4  & 80.96 & \textcolor{purple}{\textbf{15.84}}  & \textcolor{purple}{\textbf{8.60}}  &  \textcolor{teal}{\textbf{7.94}}   \\
    AOSG \citep{Han2024Unsupervised} & - / - / - / 96.4 & - / - / - / 77.7    & - / - / -  / 75.9 &  77.93 & 10.75 & 5.68 & 3.98     \\
    AGSCC \citep{Sun2024Image} & 98.3 / - / 95.9 / 97.6   & 78.2 / - / 68.0 / 77.9   & 77.3 / -  / 65.8 / 76.6  & \textcolor{teal}{\textbf{88.49}} & \textcolor{teal}{\textbf{14.82}} & \textcolor{teal}{\textbf{8.00}} & \textcolor{purple}{\textbf{9.54}}    \\
    AEKAN \citep{Liu2025AEKAN} & 98.7 / - / - / 98.3 & 83.8 / - / - / 84.7  & 83.1/ - / - / 83.9   & 81.60  & 13.09  & 7.00 & 3.56   \\

    \hline
\end{tabular}
\end{table*}

\par Table \ref{tbl:UMCD_evlaution} presents the performance of UMCD methods on \textsc{Bright} and UMCD benchmark datasets. To provide a baseline for reference, we also include a random guessing result, representing the performance floor under the UMCD setup. The UMCD benchmark datasets\footnote{These benchmark datasets are open access at \url{http://www-labs.iro.umontreal.ca/~mignotte/}}  are detailed in Figure \ref{fig:UMCD_benchmark_dataset} in Appendix \ref{app:UMCD}. Although these methods achieved considerable performance on existing UMCD benchmark datasets, their performance suffers noticeable declines on \textsc{Bright}. For instance, while they achieved F1 scores between 70-85$\%$ on existing benchmarks, their F1 scores dropped to 20$\%$ on \textsc{Bright}. This dramatic performance gap underscores the limitations of current UMCD research and highlights the challenges posed by real disaster scenarios. We identify three primary reasons for this decline:

\begin{enumerate}
    \item \textbf{Limitations in traditional UMCD datasets}. The existing UMCD datasets consist of only a handful of image pairs, often depicting simple land cover changes, such as urban expansion and deforestation, with relatively low-resolution imagery and limited geographic diversity, as shown in Figure \ref{fig:UMCD_benchmark_dataset} in Appendix \ref{app:UMCD}. These datasets fail to capture the complexity and variability found in real-world disaster scenarios. In contrast, \textsc{Bright} provides thousands of VHR multimodal image pairs across different types of disasters, significantly increasing the diversity of test cases and making it a more challenging benchmark for UMCD model evaluation. 
    \item \textbf{Interference from non-disaster changes}. Unlike previous UMCD benchmarks, where general land cover changes are the primary detection target, changes such as vegetation growth, water body shifts, and urban development may introduce noise and interfere with damage detection in real-world disaster scenarios. While prior UMCD studies treat such changes as valid detection targets, \textsc{Bright} requires methods to differentiate true building structural damage from irrelevant land cover variations, posing a unique challenge to current approaches.
    \item \textbf{Problematic evaluation protocol}. Current UMCD research follows a problematic evaluation protocol where models are trained, hyperparameters are tuned, and then validated on the same dataset. This approach can actually lead to overfitting, artificially making the model look very accurate but not generalizable to real-world scenarios. In a real-world disaster scenario, there are no labeled samples available for hyperparameter tuning, which makes the current practice unrealistic. \textsc{Bright} exposes this limitation, as models now need to be trained or tuned before being tested directly on new, unseen data, requiring truly generalizable and adaptive learning strategies.
\end{enumerate}

\par The results have highlighted the challenges of applying existing UMCD models to real-world disaster scenarios. \textsc{Bright} reveals significant limitations in current methodologies and presents new opportunities for future UMCD research.

\subsection{Evaluation of unsupervised multimodal image matching methods}

\par Precise image alignment is a critical prerequisite for any multimodal EO application. \textsc{Bright} offers a unique opportunity to evaluate the performance of existing UMIM algorithms under realistic, large-scale disaster conditions. Due to the lack of real pixel-level ground truth correspondences, we adopt a proxy evaluation strategy using manually selected control points (as shown in Figure \ref{fig:feature_points_for_UMIM} in Appendix \ref{app:registration_error_estimate}) as references. These points were selected by EO experts to represent identifiable and stable features across modalities. While this does not constitute absolute ground truth, using such human-verified correspondences provides a valuable reference. This allows us to assess how closely automated methods approximate human matching ability under multimodal and disaster conditions.

\begin{table}[!t]
  \renewcommand{\arraystretch}{1.25}
  \caption{Registration performance of different UMIM methods on the scene of Noto-Earthquake-2024. Note that the offsets here are in meters and are calculated based on the geo-coordinates of control points manually selected by EO experts. N/A means that the method is unable to complete the registration task.}
  \label{tbl:UMIM}
  \centering
  \begin{tabular}{l | c | c}
    \hline	
  Method &   Before (meters) &   After (meters)  \\
    \hline
    \hline
    FLSS \citep{Ye2017Robust} & \multirow{4}{*}{16.97} & 10.01  \\
    HOPC \citep{Ye2019Fast} & & 9.37  \\
    \cline{1-1} \cline{3-3}
     LNIFT \citep{Li2022LNIFT} &  & N/A \\
    SRIF \citep{Li2023Multimodal} &  & N/A  \\
    \hline
\end{tabular}
\end{table}

\par Table~\ref{tbl:UMIM} presents the quantitative results of four UMIM methods on the Noto-Earthquake-2024 scene. These methods fall into two main categories: feature-based methods (LNIFT \citep{Li2022LNIFT} and SRIF \citep{Li2023Multimodal}), which rely on sparse keypoint detection and matching, and area-based methods (FLSS \citep{Ye2017Robust} and HOPC \citep{Ye2019Fast}), which operate on local regions to estimate correspondences. A detailed description of each method is provided in Appendix~\ref{app:UMIM}. The experimental results reveal that traditional feature-based methods fail to achieve successful automatic registration. We attribute this limitation to the large spatial extent of the scenes and the drastic cross-modal differences, which make direct keypoint matching highly error-prone and unreliable.

\par In contrast, area-based methods, which rely on the similarity in a local region rather than point correspondence at the global level, can partially mitigate registration errors and achieve relatively more stable performance. However, these methods still face a critical limitation: they typically perform local matching within a fixed search window. This strategy is inherently inadequate for handling large displacements, especially in SAR imagery over mountainous or uneven terrain, where terrain-induced distortions can cause substantial pixel shifts that far exceed the local search range. Thus, automatic methods may serve as a useful auxiliary tool for preliminary matching or candidate generation, but human expertise remains indispensable for ensuring precise and reliable alignment in operational scenarios. We hope that \textsc{Bright} and its challenging real-world scenarios will inspire the development of new UMIM methods that are more robust to large-scale scene variations, sensitive to terrain-induced distortions, and ultimately capable of reducing human efforts in future operational EO-based disaster response workflows.

\section{Discussion}
\subsection{Limitation of \textsc{Bright}}
\par We begin this subsection by acknowledging that the composition of the \textsc{Bright} dataset is fundamentally shaped by practical constraints in data availability. While \textsc{Bright} represents a significant step forward in assembling a large-scale, multimodal, and globally distributed dataset for disaster response, it is important to recognize several inherent limitations. These limitations arise not only from the scarcity of open-access VHR SAR imagery, especially over disaster-affected regions, but also from the challenges of manual annotation and the uneven distribution of events. To provide a clearer picture for potential users, we summarize these constraints in four aspects below.

\begin{enumerate}
\item \textbf{Registration error}. \textsc{Bright} dataset consists of optical and SAR images covering the same locations. SAR images, in particular, can be distorted and stretched in certain areas. Despite thorough preprocessing, including manual alignment and cross-checking by multiple experts in EO data processing, minor alignment errors may persist, as Table \ref{tbl:registration_error} suggests. 

\item \textbf{Label quality}. The building polygons in \textsc{Bright} were manually annotated by expert annotators. Although manual labeling generally ensures high accuracy, minor errors in polygon boundaries are inevitable due to the complexity of building shapes and the variability in image resolution. These inaccuracies may slightly affect the performance of the models trained on \textsc{Bright}. Furthermore, experts assessed the extent of damage to buildings through visual interpretation of optical EO data. This process is susceptible to occasional misjudgments, contributing to label noise.

\item \textbf{Sample and regional imbalance}. Although \textsc{Bright} is rich in geographically diverse data, it has the problem of regional imbalance in the number of labels. Some of its events have more tiles and building numbers, thus, are more dominant in training and evaluation, e.g., Turkey-Earthquake-2023 (1,114 tiles) \emph{v.s.} Hawaii-Wildfire-2023 (65 tiles) in Table \ref{tbl:BRIGHT_info}. To address this, we used an additional event-level evaluation method. However, the effect of events, which account for a large percentage of the sample, on the model during the training phase is still not negligible. This may affect the generalizability of the trained model in real-world disaster scenarios. In addition, all the disaster events in \textsc{Bright} are located near the equator or in the northern hemisphere, with no events from the southern hemisphere. This spatial bias could potentially limit the applicability of trained models to regions with different building styles, land cover patterns, or SAR imaging characteristics prevalent in the southern part of the globe.

\item \textbf{Modality and temporal scope.} The dataset's scope is defined by two key characteristics of the available data. First, it exclusively utilizes single-polarization SAR imagery. The current version lacks the more informative multi-polarization or dense time-series SAR data, which, if available, could enable more nuanced damage characterization and long-term recovery monitoring, respectively. Second, the dataset's temporal coverage is concentrated on events from 2020 onwards. This is a direct consequence of its reliance on modern commercial VHR SAR providers (Capella Space and Umbra), whose open-data initiatives largely commenced around that time.

\end{enumerate}

\par Overall, despite these limitations, it is the first time such an open multimodal VHR dataset has been constructed for multimodal EO research with a large-scale and diverse disaster context.

\subsection{Significance of \textsc{Bright}}
\par Delays in both EO data acquisition and damage interpretation workflows often hinder timely disaster response \citep{Ye2024Leveraging}. Traditional expert-driven building damage mapping is time-consuming and not scalable. While ML and DL methods offer automated alternatives, their effectiveness remains limited by the scope and quality of available training data. Most existing open-source datasets are optical, restricting the models’ operational applicability in adverse weather and low-light conditions. As the first globally distributed multimodal dataset, \textsc{Bright} encompasses pre-event optical images and post-event SAR images. This unique combination overcomes the limitations of optical EO data by enabling models trained on \textsc{Bright} to monitor disaster-stricken areas regardless of weather conditions or daylight. Compared to existing building damage datasets, \textsc{Bright} offers several distinct characteristics: multimodal data, VHR imagery with sub-meter spatial resolution, coverage of five types of natural disasters and two human-made disasters, rich geographic diversity, and open access to the community. Due to these features, \textsc{Bright} is anticipated to serve as a benchmark for many future studies and practical disaster relief applications.

\par Beyond building damage assessment, \textsc{Bright} can also support several research directions within the EO and vision community. In this work, we have applied \textsc{Bright} to evaluate the performance of several UDA and SSL methods for cross-event transfer. We also demonstrated its applicability to UMCD and UMIM, showcasing its versatility as a benchmark for multiple EO challenges under real-world constraints. Furthermore, the dataset provides a strong foundation for broader multimodal EO research. Its high-quality annotations and geographic diversity directly apply to tasks, such as building footprint extraction, land cover mapping, height estimation, and EO-based visual question answering (VQA). Researchers can also repurpose or extend \textsc{Bright} to create task-specific benchmarks, enabling flexible experimentation across tasks. \textsc{Bright} is also well-positioned to support the development of EO-based foundation models, large-scale pre-trained models designed to generalize across sensors, tasks, and regions \citep{Wang2023Advancing, Hong2024SpectralGPT}. Its rich combination of modalities, spatial detail, and contextual diversity provides the data diversity required to build such general-purpose foundation models. This contribution is significant as the field moves toward creating versatile, scalable AI models that can be applied across different types of EO data and disaster scenarios \citep{LI2024Interpretable}.

\par Looking ahead, \textsc{Bright} can be further enhanced by incorporating additional modalities. For example, the inclusion of fully polarimetric SAR data would enable more nuanced damage classification than current single polarization SAR data by characterizing the different scattering properties of building materials and debris. Meanwhile, LiDAR data would offer precise 3D information to directly quantify structural collapse and enable a truly terrain-aware analysis. Future versions may also help fill current geographic gaps, including southern hemisphere regions, to ensure more globally representative coverage. Ultimately, we envision that \textsc{Bright}, true to its name, will bring even a glimmer of brightness to people in disaster-stricken areas by enabling more prompt and effective disaster response and relief.

\section{Data and Code Availability}
\par The \textsc{Bright} dataset is available at \textcolor{magenta}{\url{https://doi.org/10.5281/zenodo.14619797}} \citep{Chen2025BRIGHT}. The code for training and testing benchmark methods (including code related to IEEE GRSS DFC 2025) is accessible at \textcolor{magenta}{\url{https://github.com/ChenHongruixuan/BRIGHT}}. Models' checkpoints can be downloaded at \textcolor{magenta}{\url{https://doi.org/10.5281/zenodo.15349461}}. 

\conclusions  
\par In this paper, we introduced \textsc{Bright}, the first globally distributed multimodal dataset with open access to the community, covering 14 natural and human-made disaster events. \textsc{Bright} includes pre-event optical and post-event SAR images with sub-meter spatial resolution. Beyond introducing the dataset, we conducted a comprehensive series of experiments to validate its utility. We benchmarked several state-of-the-art supervised learning models under a standard machine learning data split. Moreover, we extended the evaluation to a cross-event transfer setting, simulating real-world scenarios where no or limited target annotations are available. Furthermore, we assessed the performance of unsupervised domain adaptation, semi-supervised learning methods, unsupervised multimodal change detection, and image matching techniques. The findings serve as performance baselines and provide valuable insights for future research in DL model design for real-world disaster response. \textsc{Bright} is an ongoing project, and we remain committed to continuously enhancing its diversity and quality by incorporating new disaster events and refining the existing data. Our objective is to improve \textsc{Bright}’s utility for practical disaster response applications at all levels (regional, national, and international) and research in the community.





\appendix

\section{Details of disaster events}\label{app:disaster_events_details}
\subsection{Explosion in Beirut, 2020}
On August 4, 2020, a massive explosion occurred at the Port of Beirut in Lebanon, caused by the improper storage of 2,750 tons of ammonium nitrate. The explosion caused widespread damage within a radius of several kilometers, significantly impacting the port and surrounding neighborhoods, including areas such as Gemmayzeh, Mar Mikhael, and Achrafieh. It resulted in 218 deaths and more than 7,000 injuries, and left approximately 300,000 people homeless \citep{fakih2021they}. Economic losses were estimated to be 15 billion USD \citep{Valsamos2021Beirut}. The disaster compounded Lebanon's ongoing economic challenges and contributed to political instability and social unrest. 


\subsection{Explosion in Bata, 2021}
On March 7, 2021, a series of four explosions occurred at the Cuartel Militar de Nkoantoma in Bata, Equatorial Guinea, caused by improperly stored explosives. The blasts led to at least 107 deaths and over 615 injuries, and widespread destruction throughout the city \citep{ocha2021bata}. A total of 243 structures were destroyed or severely damaged, displacing many residents. Around 150 families sought refuge in temporary shelters, while others stayed with relatives. Local hospitals treated more than 500 injured individuals, and the economic impact was severe, underscoring the dangers associated with the improper storage of hazardous materials.


\subsection{Volcano Eruption in DR Congo and Rwanda, 2021}
On May 22, 2021, Mount Nyiragongo in the Democratic Republic of the Congo erupted, causing widespread devastation. The eruption resulted in 32 deaths and the destruction of 1,000 homes. The displacement of thousands as lava flows threatened the city of Goma \citep{ifrc2021emergency}. Nearly 400,000 people were evacuated due to the risk of further volcanic activity, including potential magma flow beneath Goma and nearby Lake Kivu. Despite continued seismic activity, life in Goma largely returned to normal by August 2021. However, plans to relocate parts of the city remain under consideration due to the ongoing threat from the volcano.


\subsection{Earthquake in Haiti, 2021}
On August 14, 2021, a magnitude 7.2 earthquake struck Haiti’s Tiburon Peninsula, primarily affecting the Nippes, Sud, and Grand'Anse departments. The disaster caused over 2,200 deaths and more than 12,200 injuries, and left thousands homeless. The economic losses were significant, estimated at over USD 1.5 billion. Approximately 137,500 buildings, including homes, schools, and hospitals, were damaged or destroyed \citep{ocha2021haiti}. As the deadliest natural disaster of 2021, the earthquake exacerbated Haiti’s existing challenges, including widespread poverty and political instability.


\subsection{Volcano Eruption in La Palma, 2021}
On September 19, 2021, the Cumbre Vieja volcano on La Palma, part of Spain's Canary Islands, erupted following several days of seismic activity. The eruption primarily impacted the island's western side, covering over 1,000 hectares with lava and destroying more than 3,000 buildings, including the towns of Todoque and La Laguna. The lava flow, measuring about 3.5 kilometers wide and 6.2 kilometers long, reached the sea, cutting off the coastal highway and forming a new peninsula with extensive lava tubes. Although the timely evacuation of around 8,000 people prevented major casualties, one person died from inhaling toxic gases \citep{Troll2024The}. It caused significant damage to arable land and affected livelihoods, displacing thousands of residents. Economic losses exceeded 800 million Euros.


\subsection{Wildfire in Colorado, 2021}
The Marshall Fire, which started on December 30, 2021, in Boulder County, Colorado, became the most destructive wildfire in the State's history in terms of destroyed buildings. Fueled by dry grass from an unusually warm and dry season, and winds up to 185 km/h, the fire killed two people and injured eight. It destroyed 1,084 structures, including homes, a hotel, and a shopping center, causing over 2 billion USD in damage \citep{Forrister2024Analyzing}. More than 37,500 residents were evacuated, and significant damage to public drinking water systems occurred \citep{Forrister2024Analyzing}.


\subsection{Armed Conflict in Ukraine, 2022}
In February 2022, Russian forces launched a full-scale invasion of Ukraine, resulting in widespread destruction and displacement. By November 2024, total damages to Ukraine’s infrastructure had reached \$170~billion. More than 236,000 residential buildings were damaged or destroyed, including 209,000 private houses, 27,000 apartment buildings, and 600 dormitories. Over 4,000 educational institutions and 1,554 medical facilities were also affected, with extensive damage to transport, energy, and telecommunications infrastructure. The conflict has inflicted more than 40,000 civilian casualties, displaced four million people internally, and forced 6.8~millionto flee. As of late 2024, approximately 14.6~million Ukrainians require humanitarian assistance \citep{kse_damages_report_2024}.


\subsection{Earthquake in Turkey, 2023}
On February 6, 2023, a magnitude 7.8 earthquake struck southeastern Turkey near Gaziantep, followed by a magnitude 7.7 aftershock. The disaster, the most powerful earthquake in Turkey since 1939, caused widespread destruction across approximately 350,000 $km^{2}$, affecting 14 million people and displacing 1.5 million. The death toll reached 53,537, with 107,213 injuries  \citep{stl2024turkey}, making it one of the deadliest earthquakes in modern history. Economic losses were estimated at 148.8 billion USD, with over 518,000 houses and 345,000 apartments destroyed \citep{govturkiye2023recovery}. The earthquake caused severe damage to infrastructure, agriculture, and essential services, further worsening the region’s economic challenges. International aid was mobilized to support the affected populations.

\subsection{Cyclone in Myanmar, 2023}
In May 2023, Cyclone Mocha, a Category 5 hurricane, struck Myanmar, causing widespread devastation in the country's coastal regions, particularly in Rakhine State. According to official estimates, at least 148 people were killed and 132 injured, although other sources suggest higher figures \citep{worldbank2023mocha}. The cyclone affected around 1.2 million people in Rakhine alone, with over 200,000 buildings reportedly damaged or destroyed, making it one of the most destructive cyclones in the region in the past 15 years. The direct economic damage was estimated at 2.24 billion USD in damages, equivalent to 3.4\% of Myanmar's GDP \citep{worldbank2023mocha}.

\subsection{Wildfire in Hawaii, 2023}
In August 2023, a series of wildfires broke out on the island of Maui, Hawaii, causing widespread destruction and significant impacts on the local population and environment. The fires, fueled by dry conditions and strong winds, primarily affected the town of Lahaina, where at least 102 people were killed and two remain missing~\citep{hedayati20242023}. Over 2,200 buildings were destroyed, including many historic landmarks, resulting in estimated damages of 5.5 billion USD \citep{Jones2024State, ncei2025billion}. The fires prompted evacuations and led to the displacement of thousands of residents, with significant economic losses in the tourism and agriculture sectors. 


\subsection{Earthquake in Morocco, 2023}
On September 8, 2023, a 6.9 magnitude earthquake struck Morocco's Al Haouz Province near Marrakesh, causing widespread devastation. The earthquake killed nearly 3,000 people, injured more than 5,500, and displaced more than half a million \citep{ocha2023morocco}. The earthquake damaged or destroyed nearly 60,000 houses, with the heaviest losses reported in rural communities of the Atlas Mountains~\citep{IFRC2024MDRMA010OU5}.  
Overall damage is estimated at about USD 7 billion~\citep{ncei_earthquake_db}, while direct economic losses of roughly USD 30 million amount to approximately \(0.24\%\) of Morocco’s GDP. Cultural heritage in Marrakesh was also hard‑hit, sections of the UNESCO‑listed Medina and several historic mosques sustained severe damage.

\subsection{Flood and Storm in Libya, 2023}
In September 2023, Storm Daniel brought catastrophic flooding to northeastern Libya, particularly in Derna, when two dams collapsed and released an estimated 30 million cubic meters of water. At least 5,923 people were killed, though local officials warned fatalities could reach 18,000–20,000~\citep{USAIDBHA2024Libya}. Post–event analyses revealed that approximately \(10\%\) of Derna’s housing stock was destroyed and a further \(18.5\%\) sustained damage.  
Across the wider coastal belt—from Benghazi, through Jabal Al Akhdar and Al Marj, to Derna—an estimated 18,838 dwellings were damaged or obliterated~\citep{normand2024assessing}. The disaster, considered the second deadliest dam failure in history, destroyed the city’s infrastructure; four bridges collapsed, and entire neighborhoods were washed out to sea. Long-standing neglect of dam maintenance, compounded by Libya’s political turmoil, contributed significantly to the scale of the tragedy.

\subsection{Hurricane in Mexico, 2023}
In September 2023, Hurricane Norma, a Category 4 hurricane, struck the western coast of Mexico, severely affecting Sinaloa and Baja California Sur. This was followed by Hurricane Otis in October, which made landfall near Acapulco as a Category 5 hurricane. Otis was the strongest Pacific hurricane to hit Mexico, causing at least 52 deaths and leaving 32 missing. The storm caused unprecedented destruction, with more than 51,864 homes destroyed and damages estimated at 12-16 billion USD, surpassing Hurricane Wilma as the costliest Mexican hurricane~\citep{Reinhart2024OtisTCR}.

\subsection{Earthquake in Noto, 2024}
On January 1, 2024, a magnitude 7.5 earthquake struck the Noto Peninsula in Ishikawa Prefecture, Japan, reaching a maximum JMA seismic intensity of Shindo 7. The earthquake caused widespread destruction, particularly in the towns of Suzu, Wajima, Noto, and Anamizu, and triggered a 7.45-meter tsunami along the coast of the Sea of Japan. The disaster resulted in 572 deaths and over 1,300 injuries, with 193,529 structures damaged across nine prefectures~\citep{FDMA2025Noto117}. It was the deadliest earthquake in Japan since the 2011 Tōhoku disaster. As of February 20 2024, 12,929 people remained in 521 evacuation centers~\citep{JRCS2024NotoUpdate30}. The event prompted Japan's first major tsunami warning since 2011. Total damage is estimated at 16.9 trillion Japanese Yen, reflecting substantial economic losses with significant impacts on infrastructure and communities across the affected regions~\citep{Kaneko2024PrioritizingRecovery}.

\section{Feature visualization}\label{app:feat_visualization}
\par For the feature visualisation of buildings in different events in Figure \ref{fig:BRIGHT_stat_info}-(c), we specifically employ DINOv2 \citep{oquab2024dinov} to extract the high-dimensional features of buildings in different events. The corresponding reference maps are used to exclude background features. The features are then visualized using the t-SNE algorithm \citep{Laurens2008Visualizing}.

\par To visualize the spatial focus of models during building damage mapping, we apply Grad-CAM technique\footnote{https://github.com/jacobgil/pytorch-grad-cam}, a gradient-based technique for generating class-specific activation maps \citep{Selvaraju2017Grad}. We perform Grad-CAM analysis on the ChangeOS model. Specifically, we compute the activation responses for the “Damaged” and “Destroyed” categories from the following components: the third or fourth ResNet block of the two branches of encoder, capturing mid- to high-level modality-specific features; the fusion module of the damage classification branch in the decoder, reflecting early-stage fusion and decision-making. These activations help reveal which spatial regions the model relies on when assessing different damage levels and how it utilizes information from optical and SAR modalities differently, depending on the disaster context. 

\section{Manual registration and estimating registration errors}\label{app:registration_error_estimate}

\begin{figure*}[!t]
    \centering
    \includegraphics[width=7.0in]{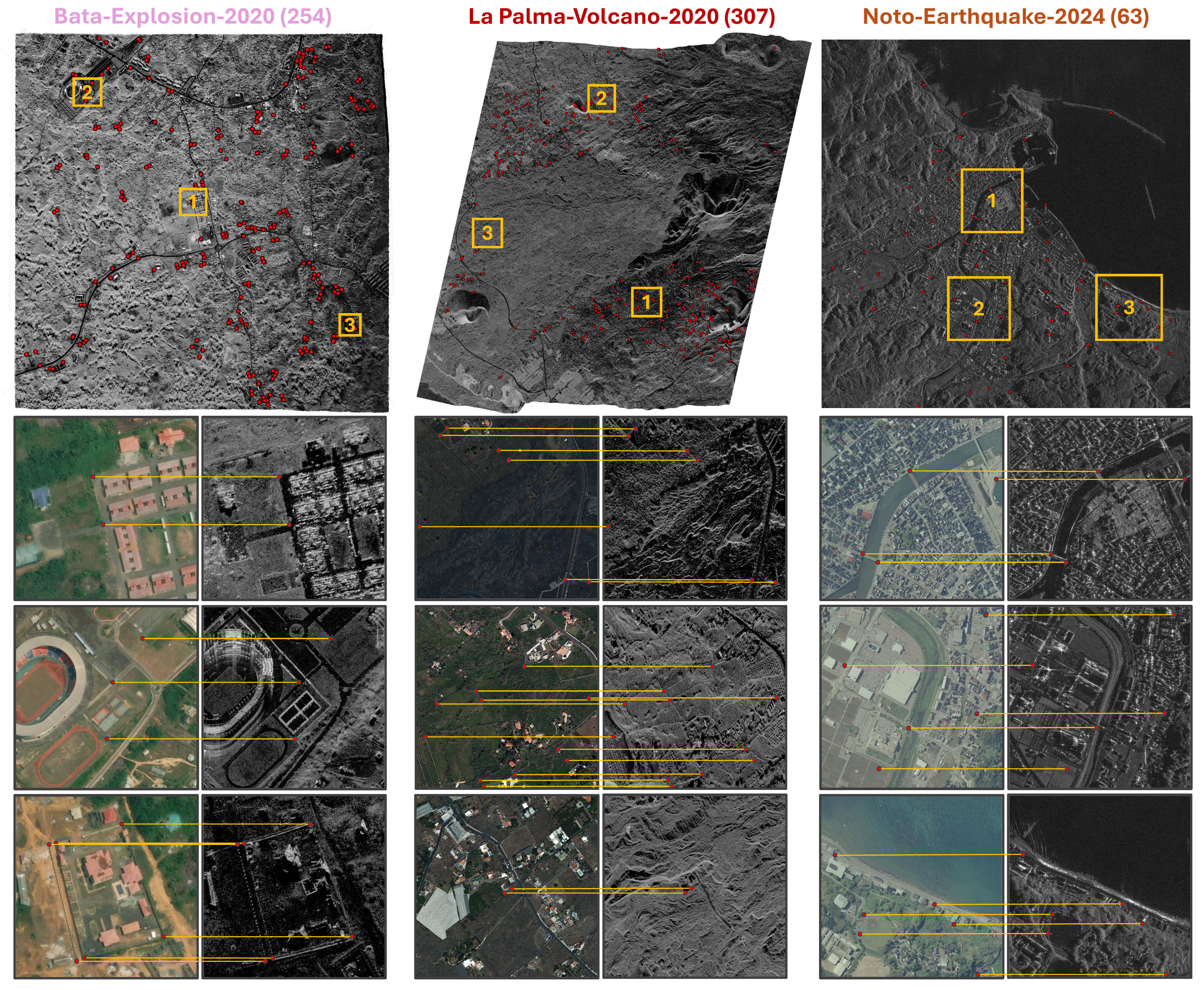}
   \caption{Overview of manually selected control points used for registration on some events. The number in parentheses of each event is the number of control points selected on that view of the image. The labeled yellow boxes highlight specific zoom-in regions that are shown in below. Data source: SAR images of Bata-Explosion-2020 and La Palma-Volcano-2020 are sourced from Capella Space Open Data Gallery while the SAR image of Noto-Earthquake-2024 is from Umbra Open Data Program. }
  \label{fig:feature_points_for_UMIM}
\end{figure*}

\par We performed the manual registration process using QGIS, with the ``Georeferencer'' plugin to align SAR images to the optical imagery as the reference. The transformation type was set to ``Thin Plate Spline'', and ``Lanczos resampling (6×6 kernels)'' was applied to achieve high-quality interpolation. The manually selected control points by EO experts on some disaster scenes are shown in Figure \ref{fig:feature_points_for_UMIM}. 

\par Despite multiple rounds of meticulous registration and cross-validation by several EO experts, registration errors between optical and SAR images cannot be completely eliminated. Since we cannot directly obtain ground truth (i.e., actual ground control points), we propose a proxy method for registration error estimation:
\begin{enumerate}
    \item Feature point selection: Select many feature points (e.g., 3000 points) from both the optical and SAR tiles using a keypoint detection algorithm.
    \item Feature extraction: Use validated and well-performing multimodal registration descriptors \citep{Li2020RIFT, Li2022LNIFT, Li2023Multimodal} to extract modality-independent features for the selected points.
    \item Feature matching: Perform feature matching between the extracted feature points across the optical and SAR images.
    \item Registration error computation: After completing the matching process, compute the pixel distances between the registered point pairs. These distances serve as a proxy for registration error.
    \item Outlier removal: To improve the reliability of the estimation, we exclude matched points with a root mean square error (RMSE) greater than a certain threshold (e.g., 20 pixels). This threshold is set because after multiple rounds of expert corrections and cross-checking, large registration errors have already been eliminated. Points exceeding this threshold can be considered outliers.
\end{enumerate}

\begin{table}[!t]
  \renewcommand{\arraystretch}{1.25}
  \caption{The estimated proxy registration error (in pixel) for each event.}
  \label{tbl:registration_error_each_event}
  \centering
  \begin{tabular}{l | c c}
    \hline	
  Event &   RMSE (All) &   RMSE (Building)  \\
    \hline
    \hline
    Beirut-EP-2020  & 1.365 &  1.365 \\
    Bata-EP-2021  & 1.226 & 1.160  \\
    Goma-VE-2021  & 0.966 & 0.975  \\
    Les Cayes-EQ-2021  & 1.392 & 1.377 \\
    La Palma-VE-2021  & 0.976 & 0.944 \\
    Marshall-WF-2021 & 1.263 & 1.266  \\
    Ukraine-AC-2022 & 0.899 &  0.882 \\
    Turkey-EQ-2023  & 0.937 & 0.901 \\
    Kyaukpyu-CC-2023  & 0.970 & 0.998  \\
    Hawaii-WF-2023  & 1.124 &  1.124 \\
    Morocco-EQ-2023  & 1.088 & 1.050  \\
    Derna-FL-2023  & 1.267& 1.255 \\
    Acapulco-HC-2023  & 1.166 & 1.148  \\
    Noto-EQ-2024  & 1.072 &  1.043 \\
    \hline
\end{tabular}
\end{table}

\par By following this method, we can estimate the registration errors between optical and SAR images in the dataset, providing a reliable foundation for further damage assessment analysis.

\section{Details of benchmark deep learning models}

\par Here, we present some implementation details of our baseline model. 
\begin{itemize}
    \item For UNet, the encoder has five convolutional blocks, each consisting of two 3$\times$3 convolutional layers and two batch normalization layers. The number of channels in the five blocks are 64, 128, 256, 512, and 1024, respectively. The decoder contains four convolutional blocks, and the number of channels is set accordingly to the number of channels of the skip-connected features extracted by the encoder.

    \item For DeepLabV3+, we applied ResNet-50 initialized with ImageNet pretraining weights \citep{He2016} as the encoder. We then modified the input channel of the encoder so that the network can predict building damage maps directly from the optical and SAR stacked images.
    
    \item For SiamAttnUNet, we followed the settings in its original literature \citep{ADRIANO2021132}. 

    \item For SiamCRNN, we applied pseudo-siamese ResNet-18 as encoder and four ConvLSTM layers \citep{Shi2015Convolutional} with 3$\times$3 convolutional kernel size and a hidden dimension of 128, as the decoder. 
    
    \item For ChangeOS, we used the ResNet-18 initialized with ImageNet pretraining weights as the encoder. We modified the encoder to a pseudo-siamese structure to extract features from pre-event and post-event images with different modalities.
    
    \item For DamageFormer, we used pseudo-siamese Swin-Transformer-Tiny \citep{Liu2021Swin} as the encoder instead of pure-siamese MixFormer \citep{Xie2021SegFormer} as in the original literature \citep{Chen2022Dual}. 
    
    \item For the ChangeMamba family of models \citep{Chen2024ChangeMamba}, we chose one of them, MambaBDA-Tiny, for our experiments. We modified the encoder to a pseudo-siamese structure to extract features from pre-event and post-event images with different modalities. 
\end{itemize}

\section{Details of post-processing methods}\label{app:pps}
\par To further refine the raw predictions from DL models and enhance the quality of building damage maps, we apply three post-processing techniques: test-time augmentation (TTA), object-based majority voting, and model ensembling in Section \ref{sec:pps}. This section provides an overview of these methods and details the specific implementation used in our study.

\begin{itemize}
    \item Test-time augmentation is a widely used technique to improve model robustness by applying transformations to the input images at inference time. Instead of making a single prediction per image, multiple augmented versions of the same input are passed through the network, and the resulting predictions are aggregated. This reduces model sensitivity to spatial variations and increases prediction stability. In our study, we apply rotation (90°, 180°, 270°) and horizontal/vertical flipping to the input images during inference. The network produces a set of predictions for each augmented version, which are then aggregated at the logit level by summing the outputs before applying the softmax function. 
    
    \item Object-based majority voting aims to enforce spatial consistency by considering entire building instances rather than making independent pixel-wise predictions. Instead of classifying each pixel separately, the final label is determined based on a majority vote across the entire building object, leading to more coherent and reliable results. We follow the setup from \citep{ZHENG2021Building}, where each building is treated as an independent object, and a weighted majority voting scheme is applied. The weighting is determined by the inverse of class proportions in the training set, meaning that underrepresented damage categories (e.g., destroyed buildings) have higher voting weights compared to dominant classes (e.g., intact buildings). 
    
    \item Model ensembling is a technique that combines predictions from multiple models to reduce uncertainty and improve generalization. By leveraging diverse models with different architectures, ensembling helps smooth out individual model biases and enhances the overall robustness of predictions. We perform ensembling by combining the results of three top-performing models: ChangeOS \citep{ZHENG2021Building}, DamageFormer \citep{Chen2022Dual}, and ChangeMamba \citep{Chen2024ChangeMamba}. Each model independently generates a damage proxy map, and the final prediction is obtained by averaging their logits before applying the softmax function. 
\end{itemize}

\begin{table*}[!t]
  \renewcommand{\arraystretch}{1.25}
  \caption{The mIoU on different events for different UDA methods in Table \ref{tbl:UDA_methods}. }
  \label{tbl:detailed_mIoU_UDA}
  \centering
  \begin{tabular}{l | c c c c c}
    \hline	
   Events & AdaptSeg  & AdvEnt  & CLAN  & PyCDA   & FDA  \\
    \hline
    \hline
     Beirut-EP-2020  & 35.55 & 33.07  & 37.25 &  32.15 & 35.89 \\ 
     Bata-EP-2021    & 37.54  & 35.20 & 36.98 &  30.01 & 36.31    \\ 
     Goma-VE-2021   & 52.49 & 51.54 & 56.91 &  49.67 & 52.46   \\ 
     Les Cayes-EQ-2021 & 37.11 & 36.86 & 39.47 & 33.40 & 33.90  \\ 
     La Palma-VE-2021   & 30.54 &  30.04 & 30.10 & 32.76 & 30.50    \\ 
     Marshall-WF-2021   & 37.20 & 35.01 & 34.35 & 30.14  & 33.93 \\ 
     Ukraine-AC-2022   & 32.36 & 32.65 & 32.21 & 32.92 &  35.03    \\ 
     Turkey-EQ-2023   & 33.69  & 34.54 & 36.68 & 34.58 & 36.82   \\ 
     Kyaukpyu-CC-2023   & 34.62 & 29.22 & 30.83 & 31.31 & 36.71 \\ 
     Hawaii-WF-2023   & 36.52 & 42.17 & 33.36 & 34.30  & 36.19  \\ 
     Morocco-EQ-2023   & 37.93 & 37.42 & 39.04 & 30.55 &  38.76    \\ 
     Derna-FL-2023   &  36.69 & 35.31 & 35.24 &  37.31 & 38.53    \\ 
     Acapulco-HC-2023 & 27.05 & 27.32 & 27.73 &  30.11 & 31.68   \\ 
     Noto-EQ-2024  & 35.49 & 35.65 &  33.99 &  25.44 & 31.39   \\ 
    \hline
    Average   & 36.05 & 35.43 & 36.01 & 33.19  &  36.29  \\ 
    \hline
\end{tabular}
\end{table*}

\section{Details of unsupervised domain adaptation methods}\label{app:UDA}

\par To evaluate the effectiveness of UDA techniques in the cross-event zero-shot setting, we selected several representative methods from the computer vision literature originally designed for semantic segmentation tasks. The following methods are implemented:
\begin{itemize}
    \item AdaptSeg~\citep{Tsai2018Learning}: an adversarial training framework that aligns output space distributions between source and target domains.

    \item AdvEnt~\citep{Vu2019ADVENT}: a variant of adversarial adaptation focusing on entropy minimization and structured output alignment.

    \item CLAN~\citep{Luo2019Taking}: a category-level alignment method that selectively aligns class-specific features.

    \item PyCDA~\citep{Lian2019Constructing}: a curriculum domain adaptation framework that iteratively refines pseudo-labels across domains.

    \item FDA~\citep{Yang2020FDA}: a simple style-transfer-based approach that transfers source images into the target domain appearance via frequency mixing.

\end{itemize}

\par All methods are implemented using DeepLabV3+ as the backbone architecture. This choice is motivated by the fact that DeepLab-based networks are commonly used in the official implementations of these UDA methods, and many codebases are tightly coupled with DeepLab’s structure. Therefore, adopting DeepLabV3+ as the shared backbone ensures compatibility with existing implementations and avoids the need for extensive code modification. It also provides a fair and consistent basis for comparison across all methods.

\par In our experimental setup, each test disaster event is treated as the target domain, and only its unlabeled optical-SAR image pairs are used for adaptation. Importantly, no labeled target domain data is used for training or model selection, in line with our zero-shot assumption. This setup is fundamentally different from conventional UDA protocols in the computer vision community, where a target domain validation set is typically available to select the best model checkpoint, and the target test set is held out purely for evaluation. In contrast, our setting is designed to reflect real-world disaster response scenarios, where it is realistic to obtain unlabeled imagery, but no ground-truth annotations are available at adaptation time. Therefore, model selection must rely solely on source domain feedback, which introduces additional challenges and more closely aligns with the operational constraints of emergency response. 

\par The hyperparameters for each method closely follow the original papers to preserve method fidelity. Below are additional implementation details specific to certain methods:
\begin{itemize}
    \item FDA: For each training iteration, we generate frequency-transferred source images using FDA and mix them 1:1 with the original source images. The combined dataset is used to train the model from scratch.

    \item PyCDA: Training proceeds in two stages. First, the model is trained on the source domain for 30,000 iterations. In the second stage, the adaptation module is applied to the target domain using pseudo-labeling and uncertainty-aware refinement as proposed in the original work.
\end{itemize}

\par For all methods, the same data preprocessing pipeline and training schedule (e.g., learning rate, batch size) are used unless otherwise specified in the original method, like those supervised models. The final evaluation is conducted on the target domain, and performance is reported in terms of mIoU.

\section{Details of semi-supervised learning methods}\label{app:SSL}

\begin{table}[!t]
  \renewcommand{\arraystretch}{1.25}
  \caption{The mIoU on different events for different semi-supervised learning methods in Table \ref{tbl:SSL_methods}.}
  \label{tbl:detailed_mIoU_SSL}
  \centering
  \begin{tabular}{l | c c c c }
    \hline	
   Events & MT  & CCT  & GCT  & CPS    \\
    \hline
    \hline
     Beirut-EP-2020  & 32.91 & 35.68 & 35.48 & 35.46       \\ 
     Bata-EP-2021    & 37.94 & 36.61 & 38.25 & 36.96       \\ 
     Goma-VE-2021   & 56.46 & 57.68 & 53.52 & 47.49      \\ 
     Les Cayes-EQ-2021    & 40.20 & 39.57 & 38.83 & 39.41     \\ 
     La Palma-VE-2021   & 33.52 & 33.03 & 36.90 &  33.29    \\ 
     Marshall-WF-2021   & 39.73 & 40.12 & 38.74 &   32.94   \\ 
     Ukraine-AC-2022   & 39.11 & 36.34& 36.44 & 31.34  \\ 
     Turkey-EQ-2023   & 38.96 & 39.72 & 36.50 & 37.15 \\ 
     Kyaukpyu-CC-2023   & 36.72 & 35.84 & 33.56 &  30.22  \\ 
     Hawaii-WF-2023   & 54.10 & 48.93 & 49.99 & 47.64  \\ 
     Morocco-EQ-2023   & 37.18 & 40.54 & 41.10 & 37.54     \\ 
     Derna-FL-2023   & 37.55 & 41.07 & 40.45 & 40.28     \\ 
     Acapulco-HC-2023 & 32.30 & 30.16 & 33.36 & 31.51   \\ 
     Noto-EQ-2024  & 43.40 & 37.61 & 37.74 & 37.65    \\ 
    \hline
    Average   & 40.00 & 39.49 & 39.34 &  37.06 \\ 
    \hline
\end{tabular}
\end{table}

\par In the one-shot cross-event transfer setting, we evaluate several representative SSL methods for semantic segmentation. These methods were originally developed for natural image domains and are adapted here to the multimodal EO-based building damage assessment scenario.

\begin{itemize}
    \item Mean Teacher (MT) \citep{Guyon2017Mean} maintains an exponential moving average (EMA) of model weights to generate stable predictions for unlabeled data, enforcing consistency between the student and teacher outputs.
    \item CCT \citep{Ouali2020Semi} applies perturbations to the input data and enforces consistency between different views of the same image using multiple decoder branches.
    \item GCT \citep{Ke2020Guided} further enhances this idea by introducing guidance from a pre-trained model to regularize learning.
    \item CPS \citep{Chen2021Semi} trains two separate networks that generate pseudo-labels for each other, encouraging cross-supervised learning on unlabeled data.
\end{itemize}

\par All methods also use DeepLabV3+ as the backbone network. This decision is consistent with their original implementations and allows for seamless integration with official codebases without the need for architectural modifications. It also ensures a fair and consistent comparison across methods.

\par In the setup, 13 labeled disaster events serve as the training and validation sets, while a single labeled sample from the test (target) event is provided to simulate a realistic one-shot adaptation scenario. In addition, the remaining unlabeled samples from the test event are made available and used to facilitate SSL.

\par Each model is trained in two stages. In the first stage, the model is trained for 10,000 iterations using only standard supervised losses, including cross-entropy and Lovász-Softmax on the labeled data from the source domain and the single labeled sample from the target domain. This stage serves to initialize the model with a stable representation before semi-supervised training begins. In the second stage, each SSL method applies its respective semi-supervised objective using the unlabeled data from the target domain. These include consistency regularization (e.g., Mean Teacher), cross-view perturbation training (e.g., CCT and GCT), or dual-network pseudo supervision (e.g., CPS). Hyperparameters for each method follow the configurations suggested in their original publications.

\par Final performance is evaluated on the target disaster event using mIoU as the primary evaluation metric.

\section{Unsupervised multimodal change detection: methods and evaluation protocol}\label{app:UMCD}
\begin{figure*}[!t]
    \centering
    \includegraphics[width=6.5in]{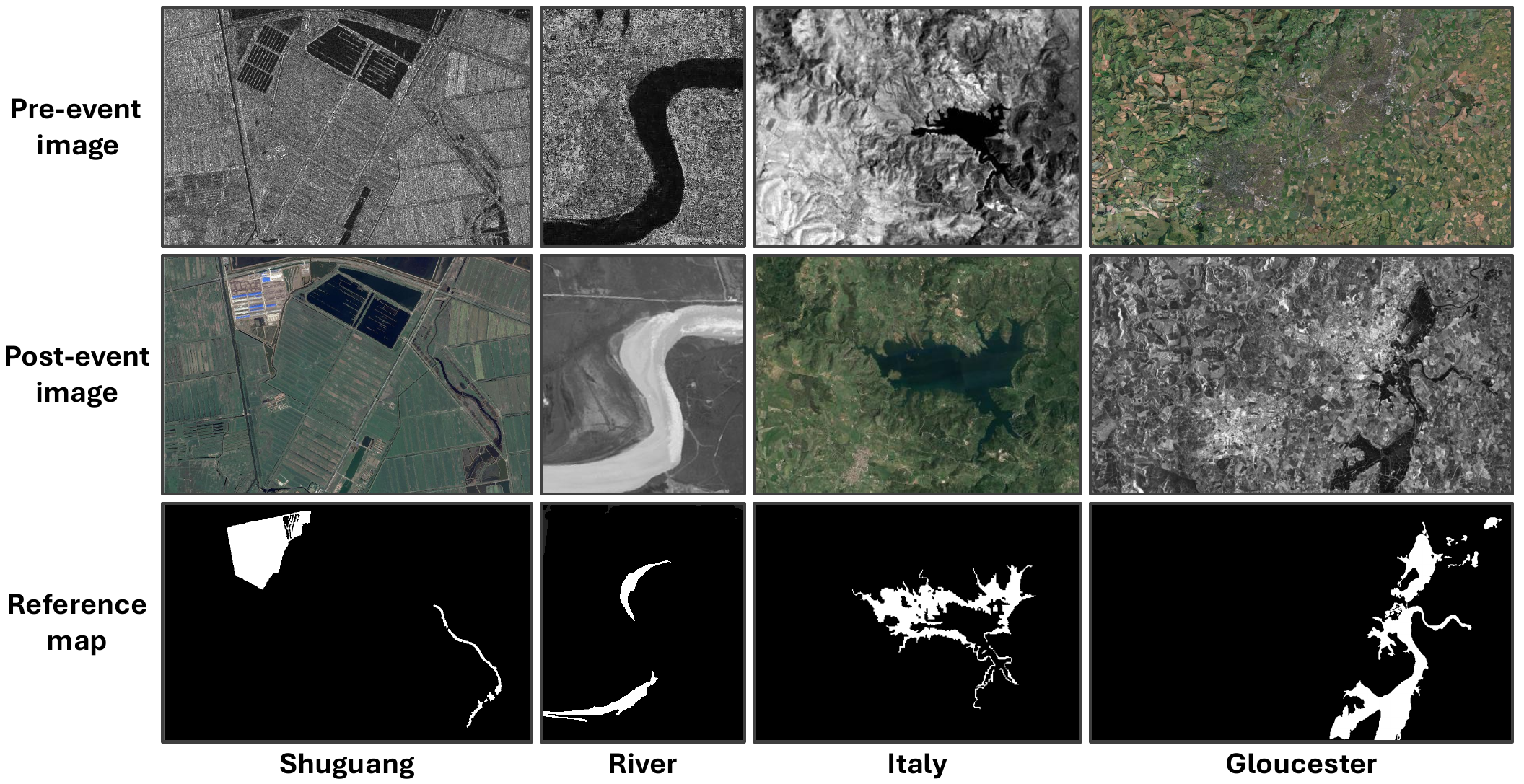}
   \caption{Illustration of benchmark datasets commonly used in unsupervised multimodal change detection. The order from left to right is the corresponding order in the Table \ref{tbl:UMCD_evlaution}. In reference maps, white color indicates changed pixels, and black color indicates unchanged pixels. Shuguang, River, and Italy datasets are sourced from \citep{Zhang2016Change}, while the Gloucester dataset is from \citep{Prendes2015Change}. }
  \label{fig:UMCD_benchmark_dataset}
\end{figure*}

\subsection{Evaluated UMCD methods}

\par We further evaluate several recent UMCD methods on \textsc{Bright}. These methods are designed to detect binary changes from paired multimodal imagery without relying on manual labels.

\begin{itemize}
    \item \emph{Random guessing} assumes each pixel has an equal probability (50\%) of being classified as changed or unchanged. This baseline provides a reference performance floor for evaluating the relative effectiveness of other UMCD methods.
    \item IRG-McS~\citep{Sun2021a} constructs superpixel-based graphs for each modality and iteratively matches their structural relationships to detect changes across heterogeneous image pairs, followed by Markov co-segmentation for final change map generation.
    \item SR-GCAE~\citep{chen2022unsupervised} learns robust graph-based representations of local and nonlocal structural relationships in multimodal images using a graph convolutional autoencoder, enabling effective change detection without supervision.
    \item FD-MCD~\citep{Chen2023Fourier} proposes a Fourier domain framework that analyzes both local and nonlocal structural relationships using graph spectral convolution and adaptive fusion, enabling robust unsupervised multimodal change detection.
    \item AOSG~\citep{Han2024Unsupervised} constructs an adaptively optimized structured graph to capture patch-level structural features in multimodal images, iteratively refining change intensity measures by fusing self-change and cross-domain structural differences for accurate unsupervised change detection.
    \item AGSCC~\citep{Sun2024Image} translates multimodal images via a structure cycle-consistent image regression framework that enforces similarity in structural graphs across domains, using adaptive graphs and multiple regularization terms to robustly detect changes without supervision.
    \item AEKAN~\citep{Liu2025AEKAN} utilizes a superpixel-based Siamese AutoEncoder built on Kolmogorov–Arnold Networks (KAN) to extract latent commonality features between multimodal images, using reconstruction and hierarchical consistency losses to detect changes in an unsupervised manner.
\end{itemize}

\subsection{A more practical evaluation protocol on \textsc{Bright}}

\par The training and evaluation protocols commonly adopted in the UMCD literature are often limited to individual image pairs (shown in Figure \ref{fig:UMCD_benchmark_dataset}), using the same scene for representation learning, hyperparameter tuning, and evaluation. While this is understandable due to the lack of large-scale public datasets in this field, such a setup fails to reflect real-world use cases and often leads to overfitting and overestimated performance. To address this, we introduce a standardized and practical evaluation protocol using the \textsc{Bright} dataset. Specifically, we use the validation set from the standard ML split in Section \ref{sec:dataset_split} as the training set for UMCD methods, including any hyperparameter or threshold tuning. These models are then evaluated on the combined training and test sets, which are strictly held out during adaptation. This avoids overlap between tuning and evaluation scenes and better simulates the real-world deployment setting, where a model is expected to generalize to new, unseen data. 

\par For evaluation, it is important to note that most of these methods were originally developed for land cover change detection, where changes correspond to transitions between semantic categories such as vegetation, water, or built-up areas. Applying them directly to disaster damage detection poses certain challenges. In particular, “Damaged” buildings do not always exhibit strong spatial or spectral signals, making them hard to distinguish in an unsupervised setting. However, “Destroyed” buildings often result in a complete change of land cover appearance (e.g., collapsed structures, debris, or ground exposure), which aligns better with the assumptions of existing UMCD methods. Therefore, in our evaluation, we treat buildings labeled as “Destroyed” in \textsc{Bright} as the positive (changed) class, and all other regions (including intact, damaged, and background) as the unchanged class. This ensures fair adaptation to disaster scenarios while respecting the original design of these methods.

\section{Details of unsupervised multimodal image matching methods}\label{app:UMIM}
\par In our extended evaluation of UMIM methods, we categorized the selected algorithms into two groups based on their matching strategies: feature-based and area-based methods. 

\par Feature-based methods, such as LNIFT~\citep{Li2022LNIFT} and SRIF~\citep{Li2023Multimodal}, aim to detect sparse keypoints independently in both optical and SAR images and compute modality-invariant feature descriptors. The matching is then performed globally by comparing descriptor similarity between the two modalities. These methods are implemented using publicly available code\footnote{https://github.com/LJY-RS/SRIF}. 

\par In contrast, area-based methods, including FLSS~\citep{Ye2017Robust} and HOPC~\citep{Ye2019Fast}, begin by identifying keypoints only in the optical image. These points are then matched by sliding a template window across a local search region in the SAR image, relying on local patch similarity. In our experiments, we used a template size of 120$\times$120 pixels, and a search window of 200$\times$200 pixels for both methods.  This category is also implemented using open-source code\footnote{https://github.com/yeyuanxin110/CFOG}. 

\par We performed registration on the entire original EO image pair from the Noto-Earthquake-2024 event. Because of the large spatial size of the input images, images were downsampled to half their original resolution to improve computational efficiency and maintain feasibility for all methods.

\par For quantitative evaluation, we adopted a control-point-based proxy metric. A set of manually selected control points, validated by EO experts shown in Figure \ref{fig:feature_points_for_UMIM}, is used to represent stable, cross-modal, and clearly identifiable features (e.g., building corners, road intersections). The average spatial offset is computed using their coordinates in a common projected coordinate system (e.g., UTM), both before and after registration. Before registration, the offset can be calculated as:
\begin{equation}
E_{\text{before}} = \frac{1}{N} \sum_{i=1}^{N} \sqrt{(x_i^{\text{opt}} - x_i^{\text{sar}})^2 + (y_i^{\text{opt}} - y_i^{\text{sar}})^2}.
\end{equation}
where $(x_i^{\text{opt}}, y_i^{\text{opt}})$ and 
$(x_i^{\text{sar}}, y_i^{\text{sar}})$ are the projected coordinates of the $i$-th control point in the optical and SAR images, respectively.

\par After registration, SAR points are transformed using the estimated mapping 
$\mathcal{T}$, and the post-registration error is computed as:
\begin{equation}
E_{\text{after}} = \frac{1}{N} \sum_{i=1}^{N} \sqrt{(x_i^{\text{opt}} - \hat{x}_i^{\text{sar}})^2 + (y_i^{\text{opt}} - \hat{y}_i^{\text{sar}})^2},
\end{equation}
where $(\hat{x}_i^{\text{sar}}, \hat{y}_i^{\text{sar}})=\mathcal{T}(x_i^{\text{sar}}, y_i^{\text{sar}})$ are transformed SAR coordinates.

\par This approach provides an alternative approximation of alignment accuracy in the absence of ground truth. Although this metric is not a substitute for full correspondence maps, it offers a practical and interpretable measure of registration performance, and enables us to assess how well unsupervised methods can approach human-level matching under complex, real-world disaster conditions.

\noappendix       




\appendixfigures  

\appendixtables   


\authorcontribution{\textbf{HC}: conceptualization (lead), data curation (lead), funding acquisition (equal), methodology (lead), project administration (lead), investigation (lead), software (lead), visualization (lead), writing – original draft preparation (lead), writing – review and editing (equal). \textbf{JS}: conceptualization (support), data curation (equal), funding acquisition (support), methodology (support), investigation (support), software (support), writing – original draft preparation (equal), visualization (support). \textbf{OD}: conceptualization (support), data curation (equal), software (support), visualization (equal), writing – review and editing (support). \textbf{CBB}: conceptualization (support), data curation (support), writing – review and editing (lead). \textbf{WX}: data curation (support), project administration (support), writing – review and editing (equal). \textbf{JW}: data curation (support), writing – review and editing (support). \textbf{XS}: funding acquisition (support), writing – review and editing (support). \textbf{YW}: writing – review and editing (support). \textbf{JX}: conceptualization (support), data curation (support), supervision (support), writing – review and editing (support). \textbf{CL}\footnote{Cuiling Lan did not participate in any activities related to the acquisition, processing, utilization, or distribution of the datasets. Her access to the data was solely limited to the information presented in the current paper, similar to that of the readers. }: conceptualization (support), supervision (support), writing – linguistic refinement (equal). \textbf{KS}: conceptualization (equal), resources (support), supervision (equal), writing – review and editing (equal). \textbf{NY}: conceptualization (equal), funding acquisition (lead), resources (lead), supervision (lead), writing – review and editing (equal). } 

\competinginterests{The contact author has declared that none of the authors has any competing interests.} 


\begin{acknowledgements}
\par This work was supported in part by the JSPS, KAKENHI under Grant Number 24KJ0652 and 22H03609, the Council for Science, Technology and Innovation (CSTI), the Cross-ministerial Strategic Innovation Promotion Program (SIP), Development of a Resilient Smart Network System against Natural Disasters (Funding agency: NIED), JST, FOREST under Grant Number JPMJFR206S, Microsoft Research Asia, Next Generation AI Research Center of The University of Tokyo, and Young Researchers Exchange Programme between Japan and Switzerland under the Japanese-Swiss Science and Technology Programme. 

\par The authors would also like to give special thanks to Sarah Preston of Capella Space, Capella Space's Open Data Gallery, Maxar Open Data Program, and Umbra's Open Data Program for providing the valuable data.
\end{acknowledgements}







\bibliographystyle{copernicus}
\bibliography{BRIGHT.bib}

\end{document}